\documentclass{article}

\PassOptionsToPackage{numbers, compress}{natbib}

\usepackage[main, final]{neurips_2026}

\usepackage[utf8]{inputenc} %
\usepackage[T1]{fontenc}    %
\usepackage{hyperref}       %
\usepackage{xurl}           %
\usepackage{booktabs}       %
\usepackage{amsfonts}       %
\usepackage{nicefrac}       %
\usepackage{microtype}      %
\usepackage[table]{xcolor}  %
\usepackage{xspace}         %
\usepackage{amsmath}
\usepackage{multirow}

\usepackage{enumitem}
\usepackage{graphicx}
\usepackage{subcaption}
\usepackage[most]{tcolorbox}
\usepackage{wrapfig}
\def\eg{\emph{e.g.}}
\def\ie{\emph{i.e.}}

\newcommand{\mysubsubsec}[1]{\textbf{#1}}
\definecolor{mygray}{gray}{0.5} 
\definecolor{myblue}{RGB}{24, 141, 228}
\definecolor{gray_bg}{gray}{0.9} 
 
\colorlet{rebcolor}{black}

\definecolor{darkblue}{rgb}{0, 0.12, 0.55}
\definecolor{darkgreen}{rgb}{0, 0.55, 0.12}
\definecolor{darkred}{rgb}{0.6,0,0}
\definecolor{darkgreen}{rgb}{0,0.6,0}
\definecolor{Gray}{gray}{0.9}
\definecolor{mymauve}{rgb}{0,0.6,0}
\definecolor{mygreen}{rgb}{0,0.6,0}
\definecolor{maincolor}{rgb}{0.2,0.5,0.7}
\definecolor{citecolor}{HTML}{0071BC}
\definecolor{linkcolor}{HTML}{ED1C24}
\hypersetup{pagebackref=false,
            breaklinks=true,
            colorlinks=true,
            citecolor=citecolor,
            linkcolor=linkcolor,
            bookmarks=false}

\newcommand{\ours}{VINCIE-NExT\xspace}
\title{VINCIE-NExT: Unlocking Video Editing from Images via In-Context Modeling}

\author{Leigang Qu$^{1}$ ~~ Feng Cheng$^{2}$ ~~ Ziyan Yang$^{2}$ ~~ Bangbang Yang$^{2}$ ~~ Zhaoyang Huang$^{2}$\\
\textbf{Wei Chow}$^{1}$ ~~ \textbf{Yicong Li}$^{3}$ ~~ \textbf{Wenjie Wang}$^{3}$ ~~ \textbf{Tat-Seng Chua}$^{1}$ ~~ \textbf{Yan Zeng}$^{2}$\\[4pt]
$^{1}$National University of Singapore ~~~~~~$^2$ByteDance Seed \\
$^3$University of Science and Technology of China \\[4pt]
{\normalfont Project Page: \url{https://vincie-next.github.io/}}
}

\begin{document}

\maketitle

\begin{figure*}[h!]
  \centering
  \includegraphics[width=\textwidth]{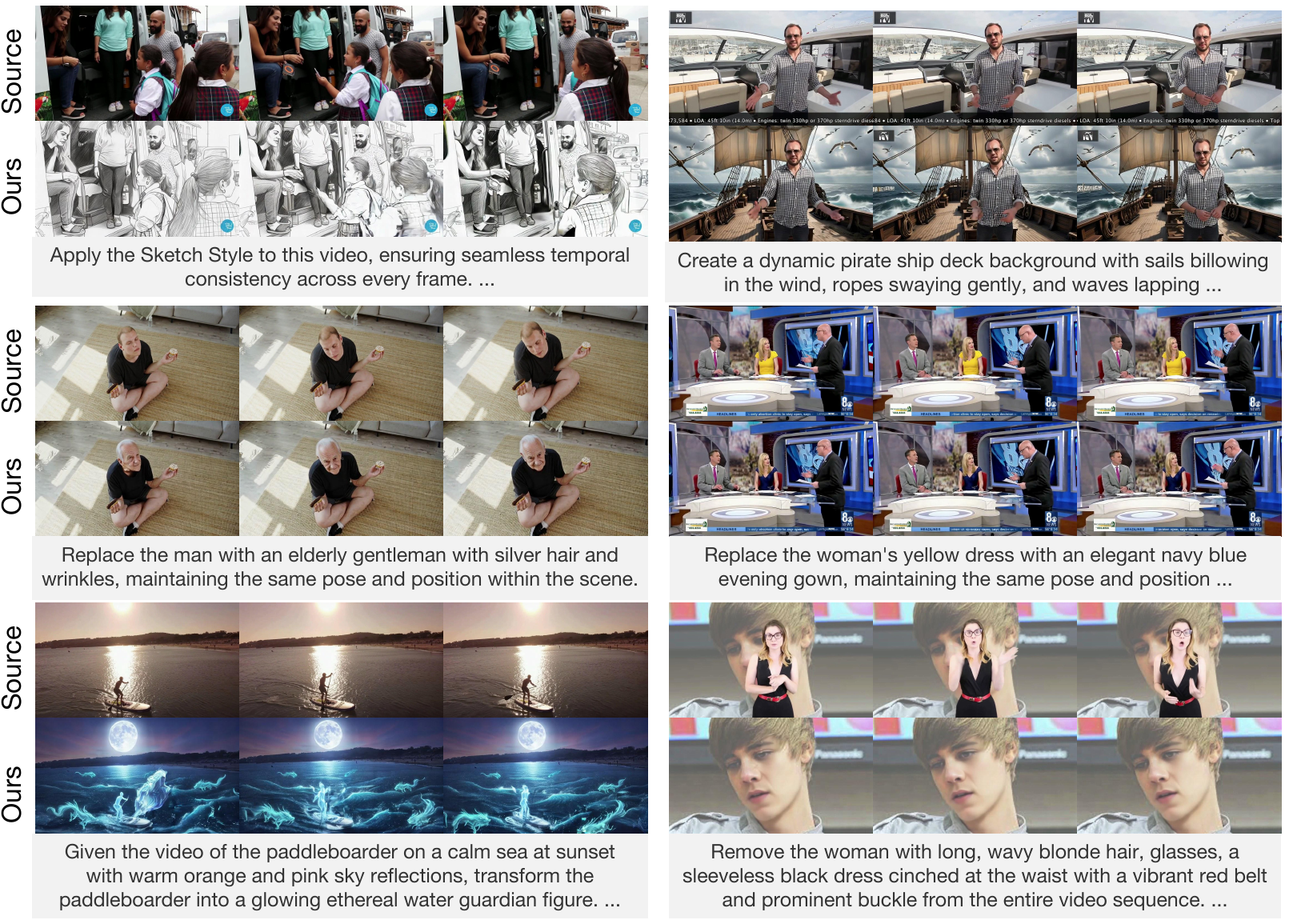}
  \caption{
      \textbf{Qualitative showcase of \ours across diverse video editing scenarios.}
      Each case presents frames extracted from the source video and the edited output. Our method enables temporally consistent stylization, dynamic background replacement, identity and attribute editing, scene transformation, and object removal, while preserving motion, structure, and visual coherence across frames.
  }
  \label{fig:teaser}
\end{figure*}

\begin{abstract}
  Building a capable video editor remains significantly harder than a video generator: editing requires (source, instruction, edited) triplets that are prohibitively expensive to annotate and difficult to synthesize at scale, whereas image editing has already reached maturity with millions of such pairs readily available.
  In this work, we introduce \textbf{\ours}, a unified framework that transfers editing capability from images to videos through in-context visual demonstrations, alleviating the need for large-scale paired video editing data.
  \ours decomposes video editing into a structured chain of composable sub-tasks ($\text{Video} \!\to\! \text{Image} \!\to\! \text{Image} \!\to\! \text{Video}$), routing editing intent through the image domain and enabling scalable joint training from heterogeneous image and video corpora under a unified diffusion objective.
  An image editing pair, synthesized by the model or supplied by the user, is prepended as an in-context visual demonstration that serves as a spatial appearance blueprint for every output frame.
  To ground appearance edits across the interleaved context, we introduce a novel position encoding that links image demonstrations and video frames in a shared spatial coordinate system, enabling pixel-faithful propagation of appearance changes to every output frame.
  Chain-of-Editing further provides principled test-time scaling: by executing the sub-task chain as progressive diffusion stages, editing quality can be improved by investing additional compute without retraining.
  Comprehensive experiments on OpenVE-Bench demonstrate the state-of-the-art performance across diverse editing categories, with ablations confirming the effectiveness of each component.
\end{abstract}

\section{Introduction}
\label{sec:intro}

Videos have become the dominant medium for creative expression, communication, and knowledge sharing, driving strong demand for video content creation. 
Recent advances in video foundation models~\cite{brooks2024sora,polyak2024moviegen,ma2025stepvideo,kling2024,seedance2026,deepmind2024veo2} have induced unprecedented visual quality~\cite{yang2024cogvideox,kong2024hunyuan,wan2025}, yet building a capable video \emph{editor} can precisely modify appearance, style, object attributes, or scene content while preserving temporal coherence, remains significantly challenging. 
\textit{The fundamental bottleneck is Data}. 
Compared with video generation which is supervised with large collections of raw video-text pairs, video editing requires \emph{pairwise} data in the form of (source video, editing instruction, edited video) triplets, which are prohibitively expensive to annotate and difficult to synthesize at scale.

Existing approaches to video editing supervision fall into two broad strategies.
The first constructs editing pairs through specialized pipelines, \eg, applying optical-flow warping, depth estimation, segmentation, or dedicated image editors frame by frame~\cite{zhang2024vivid10m,li2024editverse}.
These pipeline-based strategies suffer from compounding errors across multiple tools and limit editing diversity to the capabilities of their constituent components.
The second strategy applies pre-trained image editors independently per frame at inference time~\cite{ceylan2023pix2video,wang2023vid2vidzero}, sidestepping training-data requirements but failing to model cross-frame dependencies, leading to temporal flickering and incoherent global edits.

In recent years, \emph{image editing} has reached a level of maturity far surpassing its video counterpart.
Large-scale corpora~\cite{brooks2023ip2p, zhao2024ultraedit, wei2024omniedit, ye2025imgedit} provide millions of (source image, instruction, edited image) triplets covering diverse operations, such as style transfer, object replacement, background modification, and beyond.
Crucially, an image editing pair $(\mathbf{I}^s, \mathbf{I}^t)$ is more than a training signal: it constitutes a \emph{visual demonstration} of the desired appearance transformation.
If a generative model can leverage such a pair as in-context evidence, it may generalize the same edit to video sequences, substantially reducing the dependence of video editing on large-scale paired video data. 

This intuition drives the core design principle of our work: \emph{transferring editing capability from images to videos through in-context visual demonstrations}.
To this end, we introduce \textbf{\underline{V}}ideo \textbf{\underline{IN}}-context \textbf{\underline{C}}ha\textbf{\underline{I}}n-of-\textbf{\underline{E}}diting for next visual generation (\textbf{\ours}), a unified framework that decomposes video editing into a structured chain of composable sub-tasks, \ie, $\text{Video} \to \text{Image} \to \text{Image} \to \text{Video}$, and learns to execute this chain via joint in-context image-video modeling.
The image editing pair $(\mathbf{I}^s, \mathbf{I}^t)$, either synthesized by the model or supplied directly by users, is prepended to the interleaved context and serves as a spatial appearance blueprint that guides the generation of every frame in the edited video.

To support interleaved in-context modeling, we introduce Time Dual-Frequency 3D Rotary Position Encoding (TDF3D-RoPE), which assigns identical relative positions to spatially corresponding patches across image demonstrations and video frames. This design enables the model to precisely localize appearance changes from the image pair and propagate them faithfully to each output frame. The interleaved formulation further unlocks a two-stage inference strategy, Chain-of-Editing (\textbf{CoE}): the first stage ($\text{V}\to\text{I}\to\text{I}$) produces a high-quality edited image, and the second stage ($\text{V}\to\text{I}\to\text{I}\to\text{V}$) uses that image as a visual demonstration to guide full-video generation. 

The framework yields three key advantages.
First, sub-task decomposition enables scalable joint training from heterogeneous data: the $\text{I} \to \text{I}$ stage draws from abundant image editing pairs, and the $\text{I} \to \text{V}$ stage from image-conditioned video generation data.
Second, the in-context visual demonstration provides a rich, pixel-grounded spatial prior that guides every denoising step during generation, yielding stronger editing fidelity than text prompts alone.
Third, CoE offers principled test-time scaling, trading additional compute for higher editing quality without any retraining.
In summary, our main contributions are as follows.
\begin{itemize}[leftmargin=7.5mm]
    \setlength{\itemsep}{2pt}
    \item We propose a video editing paradigm that transfers image editing capabilities to video through in-context visual demonstrations, alleviating the need for large-scale paired video editing data.
    \item We introduce a joint in-context image--video modeling framework with a $\text{V} \to \text{I} \to \text{I} \to \text{V}$ sub-task decomposition, enabling scalable training from heterogeneous image and video corpora under a unifed diffusion learning objective.
    \item We present TDF3D-RoPE to enable pixel-faithful propagation of edits from image demonstrations to video frames, and build upon it a two-stage \emph{Chain-of-Editing} inference strategy that provides principled test-time scaling without retraining.
    \item We conduct comprehensive experiments on OpenVE-Bench~\cite{openve3m2024}, where \ours achieves state-of-the-art performance across diverse editing categories, with ablations validating the effectiveness of each component.
\end{itemize}

\section{Related Work}

\textbf{Video Generation and Editing. }
Diffusion models have become the dominant paradigm for video generation, progressing from factorized spatial-temporal attention and latent video diffusion~\cite{ho2022video,rombach2022ldm,wang2023modelscope,guo2024animatediff} to powerful Diffusion Transformer (DiT)-based generators~\cite{peebles2023scalable,brooks2024sora,yang2024cogvideox,kong2024hunyuan,polyak2024moviegen,wan2025,zheng2024opensora, qu2026ttom}.
Video editing has been addressed through inversion-based attention manipulation~\cite{mokady2023null,liu2024videop2p,qi2023fatezero,cao2023masactrl}, temporal consistency enforcement via feature propagation or flow-guided attention~\cite{geyer2024tokenflow,kara2024rave,cong2024flatten}, canonical atlas representations~\cite{kasten2021layered,bartal2022text2live,chai2023stablevideo,ouyang2024codef}, and training-based general editors operating on paired corpora or instruction-following data~\cite{molad2023dreamix,esser2023structure,wu2023tuneavideo,feng2024ccedit,wu2024fairy,ku2024anyv2v,qin2024instructvid,jiang2025vace}.
Despite this progress, training high-quality video editors remains data-intensive, motivating transfer of editing capability from the more data-rich image domain.

\textbf{Image Editing. }
Image editing with diffusion models spans a broad spectrum of approaches: training-free methods manipulate the diffusion process through attention injection or inversion~\cite{hertz2023p2p,tumanyan2023pnp}, structural conditioning adapters enable fine-grained spatial control~\cite{zhang2023controlnet}, instruction-following training on large synthetic corpora yields versatile text-driven editors~\cite{brooks2023ip2p,geng2023instructdiffusion,sheynin2024emuedit,zhao2024ultraedit,geditbench}, and reference-guided approaches support visually-grounded modifications beyond text specification~\cite{yang2023paintbyexample,wei2024omniedit}.
Most relevant to our work is the paradigm of \emph{in-context} image editing, where editing operations are demonstrated by example pairs rather than described textually~\cite{han2024incontextlora,wu2024uno,xiao2024omnigen,sun2024emu2}.
These systems demonstrate that diverse, high-quality image editing corpora can train powerful general-purpose editors—a resource we explicitly exploit to supervise video editing in this work.

\textbf{Editing Ability Transfer from Image to Video. }
A natural strategy to bootstrap video editing is to transfer capabilities from image models.
Architecture inflation—inserting temporal layers into image U-Nets or applying image-level control signals frame-by-frame~\cite{blattmann2023videoldm,blattmann2023svd,bartal2024lumiere,zhang2023controlvideo,guo2024sparsectrl}—transfers visual quality but not semantic editing behaviors.
Zero-shot methods apply image editors per-frame with temporal coherence constraints~\cite{ceylan2023pix2video,wang2023vid2vidzero,jeong2024groundavideo,yang2023rerender,yang2024fresco}, or propagate a single edited frame to the full sequence via optical flow or image-to-video priors~\cite{liang2024flowvid,ouyang2024i2vedit}; however, these approaches are prone to temporal flickering and cannot generalize to edits requiring holistic temporal reasoning.
Recent training-based efforts~\cite{qin2024instructvid,ku2024anyv2v} incorporate image editing supervision into video pipelines, yet treat the two modalities as separate tasks requiring dedicated heads or task-specific fine-tuning.
Besides, ICVE~\cite{icve2024} and EditVerse~\cite{li2024editverse} explore in-context video editing, but both perform single-pass video-to-video generation without an explicit edited keyframe.
Our approach takes a fundamentally different route: \emph{in-context image-video modeling}.
Rather than inflating architectures or applying image editors per-frame, we present image editing pairs as in-context demonstrations to a diffusion transformer that jointly attends over image and video tokens, enabling richer and more temporally consistent video editing with minimal video-specific supervision.
The resulting chain further provides a persistent image anchor, test-time extensibility, and a plug-in interface for external image editors.

\section{Methodology}\label{sec:method}

\begin{figure}[htbp]
    \centering
    \includegraphics[width=\textwidth]{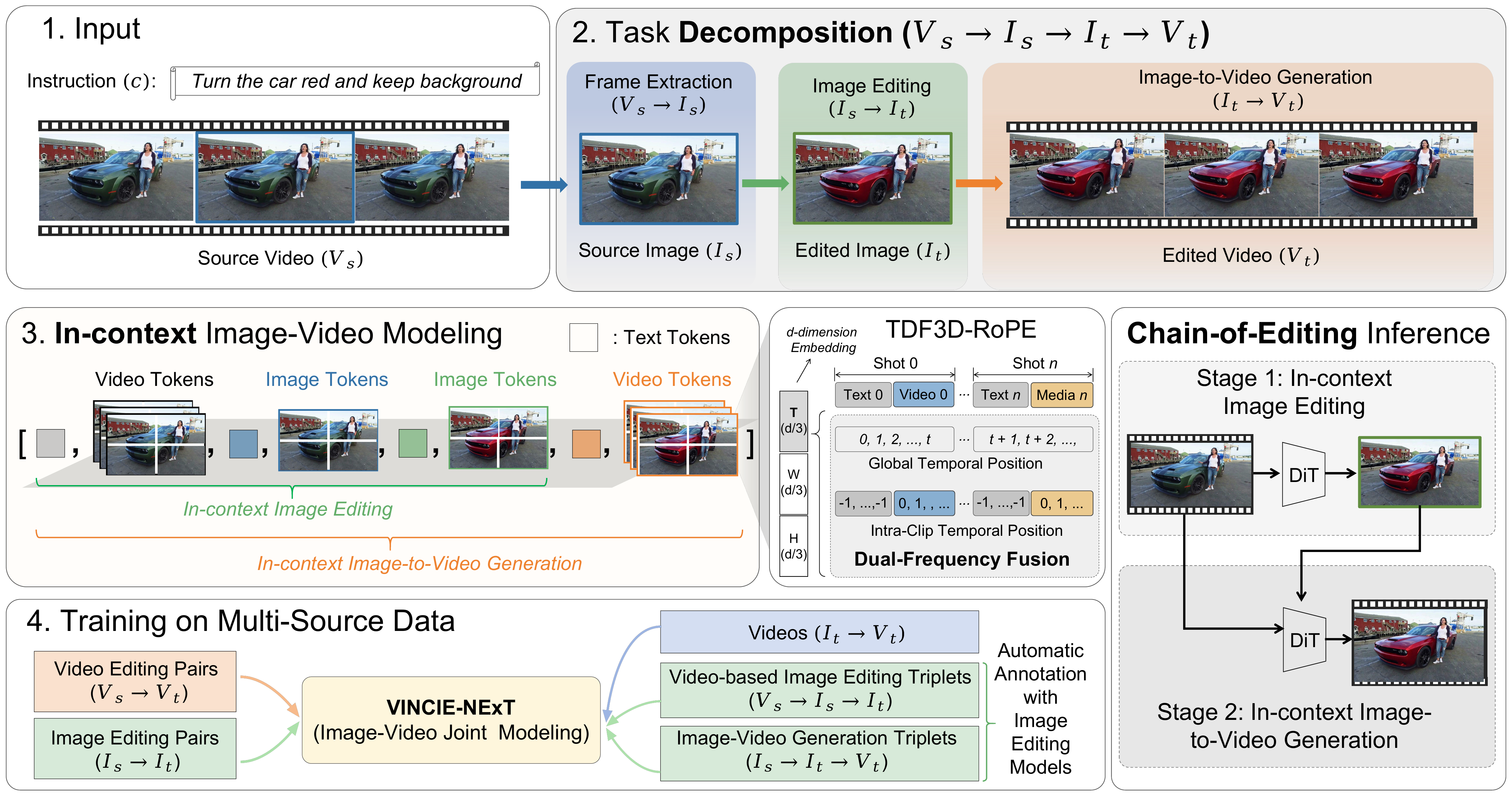}
    \caption{Overview of our proposed \ours framework. The video editing process is decomposed into a structured chain of sub-tasks ($\text{Video} \rightarrow \text{Image} \rightarrow \text{Image} \rightarrow \text{Video}$). The image editing pair $(\mathbf{I}^s, \mathbf{I}^t)$ serves as an in-context visual demonstration that transfers image editing priors to video generation via a unified diffusion backbone.}
    \label{fig:framework}
\end{figure}

We present \ours, a video editing framework that \emph{transfers editing capability from images to videos through in-context visual demonstrations}, as demonstrated in Fig.~\ref{fig:framework}.
We decompose video editing into a chain: $\text{Video} \rightarrow \text{Image} \rightarrow \text{Image} \rightarrow \text{Video}$, routing editing intent through the image domain where mature large-scale priors are readily available. An image editing pair $(\mathbf{I}^s, \mathbf{I}^t)$, synthesized or user-supplied, serves as a spatial blueprint guiding every output frame.
The framework centers on three components: (i)~\emph{Decomposition of Video Editing}, (ii)~\emph{Joint In-context Image--Video Modeling}, and (iii)~\emph{Chain-of-Editing}, a two-stage inference strategy executing $\text{V}{\to}\text{I}{\to}\text{I}$ then $\text{V}{\to}\text{I}{\to}\text{I}{\to}\text{V}$.

\subsection{Decomposition of Video Editing}

\mysubsubsec{Problem formulation.}
Let $\mathbf{V}^s = (f_1, \ldots, f_T)$ denote a source video of $T$ frames, $c$ an editing instruction, and $\mathbf{V}^t = (f_1', \ldots, f_T')$ the corresponding edited video.
Conventional video editing approaches model $p(\mathbf{V}^t \mid \mathbf{V}^s, c)$ directly, requiring the model to simultaneously handle high-level semantic editing, fine-grained appearance control, and cross-frame temporal consistency, a joint objective that demands large-scale paired video data which is costly to acquire.

\mysubsubsec{Sub-task decomposition.}
We instead factorize the editing process by introducing intermediate image representations.
Given a representative keyframe $\mathbf{I}^s$ extracted from $\mathbf{V}^s$ and its instruction-edited counterpart $\mathbf{I}^t$, the joint distribution over the full generation chain factorizes as:
\begin{equation}
    p(\mathbf{I}^s, \mathbf{I}^t, \mathbf{V}^t \mid \mathbf{V}^s, c) =
    \underbrace{p(\mathbf{I}^s \mid \mathbf{V}^s, c)}_{\text{V} \to \text{I}} \cdot
    \underbrace{p(\mathbf{I}^t \mid \mathbf{I}^s, \mathbf{V}^s, c)}_{\text{I} \to \text{I}} \cdot
    \underbrace{p(\mathbf{V}^t \mid \mathbf{I}^t, \mathbf{I}^s, \mathbf{V}^s, c)}_{\text{I} \to \text{V}}
    \label{eq:decomp}
\end{equation}
Each factor corresponds to a distinct sub-task with a clearly delineated role.
\text{V$\to$I} extracts a semantically representative image from the source video, anchoring the editing operation to a single spatial context.
\text{I$\to$I} applies the instruction-guided edit at the image level, where mature image editing priors offer fine-grained control.
\text{I$\to$V} recomposes the edited image into a temporally consistent target video, conditioned on both the edited frame and the original video structure.
At inference, the temporal mid-frame of $\mathbf{V}^s$ is chosen as $\mathbf{I}^s$, a safe default. In practice, any frame can be used, as training spans diverse temporal positions and Temporal Position Randomization (\S\ref{sec:tdf3d}) makes the model robust to this choice.

\mysubsubsec{Joint training.}
This decomposition enables scalable supervision by allowing each sub-task to draw from independently collected data: abundant image editing pairs for I$\to$I, image-conditioned video generation data for I$\to$V, and video--keyframe correspondence data for V$\to$I.
We define the combined training corpus as $\mathcal{D} = \mathcal{D}_{V \to I} \cup \mathcal{D}_{I \to I} \cup \mathcal{D}_{I \to V} \cup \mathcal{D}_{\text{full}}$, where $\mathcal{D}_{\text{full}}$ contains full-chain samples that supervise the complete pipeline.
Sub-tasks are sampled randomly during training, and all are supervised with a unified flow-matching objective over interleaved visual--text sequences:
\begin{equation}
    \mathcal{L} = \mathbb{E}_{(x,\,y)\,\sim\,\mathcal{D},\;\epsilon\,\sim\,\mathcal{N}(0,\mathbf{I}),\;\tau\,\sim\,\mathrm{LogitNormal}(0,1)}\;
    \bigl\|(\epsilon - y) - \epsilon_\theta(y_\tau,\, x,\, \tau)\bigr\|^2
    \label{eq:loss}
\end{equation}
where $y_\tau = (1-\tau)\,y + \tau\,\epsilon$ is the target visual segment noised to timestep $\tau \in [0,1]$, $\epsilon_\theta$ predicts the velocity $\epsilon - y$, and $x$ denotes the conditioning context (source video tokens, instruction, and any preceding interleaved segments).
This joint training regime allows the model to share a single set of parameters across all sub-tasks, facilitating cross-task knowledge transfer and, in particular, enabling image editing to generalize video.

\subsection{Joint In-context Image-Video Modeling}

\mysubsubsec{Unified visual tokenization.}
Images and videos are encoded by a shared video VAE, \ie,  $\mathcal{E}$, into continuous latents with $8\times$ spatial and $4\times$ temporal compression, which are then patchified with spatial patch size $p$.
A frame $f \in \mathbb{R}^{H \times W \times 3}$, encoded as a single-frame clip, yields a flat sequence of $N = \frac{H}{8p} \times \frac{W}{8p}$ tokens, and a video $\mathbf{V} = (f_1, \ldots, f_T)$ is encoded jointly into $T' = 1 + \frac{T-1}{4}$ latent frames, yielding $NT'$ tokens.

\mysubsubsec{Interleaved sequence construction.}
The full generation chain (Eq.~\ref{eq:decomp}) is flattened into one contiguous token sequence:
\begin{equation}
    \mathbf{x} = \bigl[\,c_1,\;\mathcal{E}(\mathbf{V}^s),\;c_2,\;\mathcal{E}(\mathbf{I}^s),\;c_3,\;\mathcal{E}(\mathbf{I}^t),\;c_4,\;\mathcal{E}(\mathbf{V}^t)\,\bigr]
    \label{eq:seq}
\end{equation}
where each visual segment is paired with its own text prompt: $c_1$ is a media reference tag for the source video, $c_2$ is a frame-sampling prompt for V$\to$I, $c_3$ is the editing instruction for I$\to$I, and $c_4$ is a image-to-video generation prompt for I$\to$V.
The model generates each visual segment via a diffusion process conditioned on all preceding tokens in the interleaved context.
Crucially, the image pair $\bigl(\mathcal{E}(\mathbf{I}^s),\, \mathcal{E}(\mathbf{I}^t)\bigr)$ forms an \emph{in-context visual demonstration}: when generating $\mathcal{E}(\mathbf{V}^t)$, the model has access to the full source-to-target appearance mapping at the image level as a pixel-grounded spatial prior for every frame.
For sub-task training, any contiguous sub-chain of $\mathbf{x}$ with its paired prompts (\eg, $[c_1, \mathcal{E}(\mathbf{I}^s), c_3, \mathcal{E}(\mathbf{I}^t)]$ for standalone image editing) constitutes a self-contained training sample under the loss in Eq.~\ref{eq:loss}, allowing sub-task-specific data to be mixed in without requiring the full chain.

\subsubsection{Time Dual-Frequency 3D Rotary Position Encoding (TDF3D-RoPE)}\label{sec:tdf3d}

The interleaved sequence (Eq.~\ref{eq:seq}) mixes text with heterogeneous visual segments, so naive sequential indices would conflate content identity with sequence order.
We use a factorized 3D RoPE assigning each token a position triplet $(p_t, p_h, p_w)$ with separate frequency tensors for text and visual tokens.

In-context editing imposes two conflicting positional demands: the model should (i)~distinguish tokens from different shots (\eg, $\mathbf{I}^s$ vs.\ $\mathbf{I}^t$) to maintain causal ordering, while (ii)~recognizing spatially corresponding patches across shots to propagate edits pixel-faithfully.
We resolve this by computing two independent 3D frequency tensors and fusing them via element-wise addition.
The \textbf{inter-shot component} assigns monotonically increasing global offsets across shots, making every shot globally distinguishable.
The \textbf{intra-video component} assigns raw spatial positions $(t_i, h, w)$ with zero global offset, so the same spatial location in any two shots shares identical $(p_h, p_w)$, enabling purely content-driven attention between corresponding patches and pixel-faithful edit propagation from $\mathbf{I}^s{\to}\mathbf{I}^t$ to $\mathbf{V}^t$.
Under RoPE~\cite{su2024roformer}, frequency addition encodes both objectives simultaneously within a single rotation with no modality-type embeddings.
Formally, for shot $k$ with accumulated global offset $\tau_k$, a visual token at local coordinates $(t_i, h, w)$ receives
\begin{equation}
    \mathbf{p}_{\mathrm{inter}} = (\tau_k {+} t_i,\, \tau_k {+} h,\, \tau_k {+} w), \quad
    \mathbf{p}_{\mathrm{intra}} = (t_i, h, w), \quad
    \tau_{k+1} = \tau_k + l_k + \max\bigl(\lceil t_{\mathrm{last}} \rceil, H, W\bigr),
    \label{eq:tdf3d}
\end{equation}
where $l_k$ is the text length of shot $k$, $t_{\mathrm{last}}$ is the timestamp of the last frame in shot $k$, $H, W$ is the spatial grid size, and the rotation uses the frequency-domain sum $\mathbf{p}_{\mathrm{inter}} + \mathbf{p}_{\mathrm{intra}}$.

\mysubsubsec{Temporal position randomization for image training data.}
Assigning a fixed temporal index $t_i = 0$ to all training images creates a spurious correlation between temporal position and the image modality.
We replace it with $\tilde{t} \sim \mathcal{U}[0,\, T_{\max}]$, compelling the model to ground correspondences in spatial content rather than temporal identity.
Full formulas are given in Appendix~\ref{sec:app:method}.

\subsection{Chain-of-Editing}

The decomposition in Eq.~\ref{eq:decomp} is instantiated at inference time as two sequential diffusion stages.
\emph{Chain-of-Editing} (CoE) advances one sub-task at a time, carrying the outputs of prior stages as context for all subsequent ones.

\mysubsubsec{Stage 1 — Image editing (V$\to$I$\to$I).}
The first stage generates the edited keyframe $\mathbf{I}^t$ by assembling an interleaved 3-shot context from the source video $\mathbf{V}^s$, source keyframe $\mathbf{I}^s$, and the target image slot, then running a full diffusion sampling loop to yield the denoised latent $\hat{z}^t$.

\mysubsubsec{Stage 2 — Video generation (V$\to$I$\to$I$\to$V).}
The second stage generates $\mathbf{V}^t$ conditioned on $\mathbf{V}^s$, $\mathbf{I}^s$, and $\hat{z}^t$, using a 4-shot context ($\mathbf{V}^s$, $\mathbf{I}^s$, $\hat{z}^t$, target video slot) processed by the \emph{same} backbone, which now attends jointly across frames to model cross-frame dependencies.
Critically, $\hat{z}^t$ is inserted directly in latent space, without any VAE decode and re-encode cycle, preserving the output fidelity at Stage~1 and avoiding compounding reconstruction errors.

\mysubsubsec{Per-shot timestep conditioning.}
Conditioning shots receive timestep $t = 0$ (fully denoised clean signals), while only the target shot receives the active diffusion timestep $\tau \in [0, 1]$.
The entire multi-shot sequence is passed to the DiT in a single forward call, so conditioning and target tokens attend to each other at every denoising step.

\mysubsubsec{Test-time scaling.}
Because each stage is an independent diffusion process over an explicit interleaved context, the chain can be extended at test time without retraining, \eg, iterating image editing multiple times before video generation, trading additional compute for higher fidelity.
See Appendix~\ref{sec:app:method} for implementation details of all stages.

\subsection{In-context Video-based Image Editing Data Construction}\label{sec:viie_data}

Beyond video editing pairs, we augment training with \emph{video-based in-context image editing} (V$\leftrightarrow$I$\leftrightarrow$I) data: each sample pairs an original video keyframe with an edited counterpart produced by a state-of-the-art image editor under an instruction annotated by GPT-4o (see Appendix~\ref{app:annotation}).
Unlike conventional image editing pairs trained in isolation, each sample keeps the source video attached as context, \ie, $[\mathcal{E}(\mathbf{V}^s), \mathcal{E}(\mathbf{I}^s), \mathcal{E}(\mathbf{I}^t)]$, teaching exactly the V$\to$I$\to$I sub-step whose output is reused as the demonstration for V$\to$I$\to$I$\to$V, so that image editing ability transfers into video editing rather than remaining a standalone skill.
This augmentation provides three key benefits:
\textbf{(1) Broader editing knowledge.} Diverse categories (\eg, attribute modification, expression transfer, object removal) expose the model to a wider range of editing concepts than video pairs alone.
\textbf{(2) Stronger intermediate editing.} Training explicitly on the image-editing stage of Chain-of-Edit improves the quality of the edited keyframe that ground subsequent video generation.
\textbf{(3) Scalability.} High-quality $(\mathbf{I}^s, c, \mathbf{I}^t)$ triplets, can be produced from arbitrary video frames without the video pairs that are costly to collect.

\section{Experiments}\label{sec:exp}

\subsection{Experimental Setup}

\noindent\textbf{Training Data.}
\ours is trained on three complementary data sources corresponding to the \textbf{I2I}, \textbf{V2V*}, and \textbf{V$\leftrightarrow$I$\leftrightarrow$I} splits in the ablation study (Tab.~\ref{tab:ablation}).
\textbf{I2I} uses paired image editing data from OmniEdit~\cite{wei2024omniedit}.
\textbf{V2V*} uses paired video editing data from OpenVE~\cite{openve3m2024}, where * denotes that each pair is reformatted into sub-chains of our decomposed V$\to$I$\to$I$\to$V format (\eg, V$\to$V, I$\to$I, I$\to$V) rather than used as a plain V2V pair.
\textbf{V$\leftrightarrow$I$\leftrightarrow$I} is our core novel data source: for each raw video, GPT-4o generates editing instructions that are applied per-frame by Seedream~4.5~\cite{seedream2025}, forming (source video, sampled frame, edited frame) triplets \emph{without} any paired video editing supervision.
The three sources contain 1.20M, 2.45M, and 1.63M samples, respectively.
Full construction details are in Appendix~\ref{sec:app:exp_setting}.

\noindent\textbf{Benchmark and Evaluation.}
We evaluate on OpenVE-Bench~\cite{openve3m2024}, comprising 431 video clips across eight editing categories (six spatially-aligned: Global Style, Background Change, Local Change, Local Remove, Local Add, Subtitle Edit; two non-spatially-aligned: Creative Edit, Camera Edit).
Following OpenVE-Bench, we use an MLLM-based evaluator (Gemini 2.5 Pro) that scores each output on a 1--5 scale along three dimensions: \textit{Instruction Compliance}, \textit{Consistency \& Detail Fidelity}, and \textit{Visual Quality \& Stability}.
We compare against seven open-source baselines (VACE~\cite{jiang2025vace}, OmniVideo~\cite{omnivideo2024}, InsViE~\cite{insvie2024}, Lucy-Edit~\cite{lucyedit2024}, ICVE~\cite{icve2024}, DITTO~\cite{ditto2024}, OpenVE-Edit~\cite{openve3m2024}) and the proprietary Runway Aleph.
We further compare with AnyV2V~\cite{ku2024anyv2v}, a representative image-editing-driven method, in Appendix~\ref{sec:app:anyv2v}.
Full benchmark and baseline details are in Appendix~\ref{sec:app:exp_setting}.

\noindent\textbf{Implementation Details.}
The DiT backbone $\epsilon_\theta$ is a 3B MM-DiT initialized from the same text-to-video pretrained backbone as VINCIE~\cite{qu2025vincie}, the visual tokenizer $\mathcal{E}$ is a video VAE inflated from SD3~\cite{esser2024sd3} ($\times 8$ spatial, $\times 4$ temporal downsampling, 16 channels; $p=(1,2,2)$), and the text encoder is Flan-T5~\cite{chung2024flant5}. Only the DiT is trained, and it is shared by all sub-tasks and both inference stages.
We train on 32 H100 GPUs for about 150 hours (42k steps, lr $ = 5\times10^{-5}$) in two stages on the same data: Stage~1 at $256\times256$, then Stage~2 at $640\times480$ initialized from Stage~1. Inference uses up to 65 frames at 12 FPS, 32 sampling steps, and a classifier-free guidance (CFG) scale of 2.5 (full settings in Tab.~\ref{tab:impl}). Tab.~\ref{tab:perf_gemini} reports the Stage-2 model. Due to limited compute, all ablations use the Stage-1 model with otherwise identical settings. 

\subsection{Performance Comparison}

\begin{table*}[ht]
    \centering
    \small
    \setlength{\tabcolsep}{3pt} %
    \caption{Performance comparison of video editing methods across eight editing categories, evaluated by Gemini 2.5 Pro, with automatic ratings on a 1--5 scale following the OpenVE-Bench protocol. \#Reso. denotes the output resolution. \color{mygray}{Grey} rows denote closed-source commercial models.}
    \label{tab:perf_gemini}
    \resizebox{0.95\textwidth}{!}{%
    \begin{tabular}{l|c|c|cccccccc}
    \toprule
    Methods & \#Reso. & Overall & \begin{tabular}[c]{@{}c@{}}Global\\ Style\end{tabular} & \begin{tabular}[c]{@{}c@{}}Background\\ Change\end{tabular} & \begin{tabular}[c]{@{}c@{}}Local\\ Change\end{tabular} & \begin{tabular}[c]{@{}c@{}}Local\\ Remove\end{tabular} & \begin{tabular}[c]{@{}c@{}}Local\\ Add\end{tabular} & \begin{tabular}[c]{@{}c@{}}Subtitle\\ Edit\end{tabular} & \begin{tabular}[c]{@{}c@{}}Creative\\ Edit\end{tabular} & \begin{tabular}[c]{@{}c@{}}Camera\\ Edit\end{tabular} \\ 
    \midrule
    \color{mygray}{Runway Aleph} & \color{mygray}{$1280 \times 720$} & \color{mygray}{3.65} & \color{mygray}{3.72} & \color{mygray}{2.62} & \color{mygray}{4.18} & \color{mygray}{4.16} & \color{mygray}{2.78} & \color{mygray}{3.62} & \color{mygray}{3.64} & \color{mygray}{4.53} \\
    VACE~\cite{jiang2025vace} & $1280 \times 720$ & 1.57 & 1.49 & 1.55 & 2.07 & 1.46 & 1.26 & 1.48 & 1.47 & 1.62 \\
    OmniVideo~\cite{omnivideo2024} & $640 \times 352$ & 1.31 & 1.11 & 1.18 & 1.14 & 1.14 & 1.36 & 1.00 & 2.26 & 1.00 \\
    InsViE~\cite{insvie2024} & $720 \times 480$ & 1.53 & 2.20 & 1.06 & 1.48 & 1.36 & 1.17 & 2.18 & 2.02 & 1.09 \\
    Lucy-Edit~\cite{lucyedit2024} & $1280 \times 704$ & 2.15 & 2.27 & 1.57 & \underline{3.20} & 1.75 & \textbf{2.30} & 1.61 & \underline{2.86} & 1.61 \\
    ICVE~\cite{icve2024} & $384 \times 240$ & 2.07 & 2.22 & 1.62 & 2.57 & \underline{2.51} & 1.97 & 2.09 & 2.41 & 1.11 \\
    DITTO~\cite{ditto2024} & $832 \times 480$ & 1.98 & \underline{4.01} & 1.68 & 2.03 & 1.53 & 1.41 & 2.81 & 1.23 & 1.32 \\
    OpenVE-Edit~\cite{openve3m2024} & $1280 \times 704$ & \underline{2.49} & 3.16 & \underline{2.36} & 2.98 & 1.85 & 2.15 & \underline{2.91} & 2.31 & \underline{2.02} \\
    \midrule
    \rowcolor{gray_bg}
    \textbf{Ours} & $640 \times 480$ & \textbf{3.08} & \textbf{4.17} & \textbf{2.55} & \textbf{3.48} & \textbf{3.24} & \underline{2.27} & \textbf{3.45} & \textbf{3.41} & \textbf{2.06} \\
    \bottomrule
    \end{tabular}
    }%
\end{table*}

\mysubsubsec{Overall results.}
As shown in Tab.~\ref{tab:perf_gemini}, \ours achieves an overall score of \textbf{3.08}, surpassing all open-source methods by a substantial margin, with 24\% relative gain over OpenVE-Edit, closing the gap to the proprietary Runway Aleph, while substantially reducing the dependence on large-scale paired video editing data. 
Results are consistent under Seed1.6-VL as an independent judge (Appendix~\ref{sec:app:additional_exp}).

\mysubsubsec{Category-level analysis.}
Global Style is the standout category (4.17 vs.\ Runway Aleph's 3.72): the in-context editing paradigm directly encodes the full appearance transformation, and TDF3D-RoPE propagates it temporally.
Local Remove also sees dramatic improvement (1.85$\to$3.24 over OpenVE-Edit), as routing edits through a mature image editor yields coherent inpainting.
Most other appearance-editing categories, such as Local Change, Subtitle Edit, and Creative Edit, similarly benefit from the design.
The primary limitation is Camera Edit, compared with Runway Aleph, which requires novel-viewpoint synthesis, \ie, a geometric operation outside the scope of the appearance editing. Local Add and Background Change also show room for improvement, as both demand synthesizing content absent from the source video.

\subsection{Ablation Study}\label{sec:exp:ablation}

\begin{table*}[ht]
    \centering
    \small
    \setlength{\tabcolsep}{3pt}
    \caption{Ablation study on data composition and Chain-of-Editing (CoE) on OpenVE-Bench, evaluated by Gemini 2.5 Pro.
    Rows with \colorbox{gray_bg}{grey background} indicate that CoE is applied at inference time.
    I2I, V2V*, and V$\leftrightarrow$I$\leftrightarrow$I are the three training data sources.
    Training on V$\leftrightarrow$I$\leftrightarrow$I data alone already enables meaningful video editing,
    and CoE provides consistent gains when this chain data is present.
    The full combination of all three data types with CoE achieves the best overall score.
    All rows use the model after the first training stage at $256\times256$ resolution. }
    \label{tab:ablation}
    \resizebox{0.95\textwidth}{!}{%
    \begin{tabular}{ccc|c|c|cccccccc}
    \toprule
    \multicolumn{3}{c|}{Data Composition} & \multirow{2}{*}{CoE} & \multirow{2}{*}{Overall} & \multirow{2}{*}{\begin{tabular}[c]{@{}c@{}}Global\\ Style\end{tabular}} & \multirow{2}{*}{\begin{tabular}[c]{@{}c@{}}Background\\ Change\end{tabular}} & \multirow{2}{*}{\begin{tabular}[c]{@{}c@{}}Local\\ Change\end{tabular}} & \multirow{2}{*}{\begin{tabular}[c]{@{}c@{}}Local\\ Remove\end{tabular}} & \multirow{2}{*}{\begin{tabular}[c]{@{}c@{}}Local\\ Add\end{tabular}} & \multirow{2}{*}{\begin{tabular}[c]{@{}c@{}}Subtitle\\ Edit\end{tabular}} & \multirow{2}{*}{\begin{tabular}[c]{@{}c@{}}Creative\\ Edit\end{tabular}} & \multirow{2}{*}{\begin{tabular}[c]{@{}c@{}}Camera\\ Edit\end{tabular}} \\[2pt]
    I2I & V2V* & V$\leftrightarrow$I$\leftrightarrow$I & & & & & & & & & & \\
    \midrule

    \checkmark &  &  &  & 1.14 & 1.29 & 1.00 & 1.05 & 1.05 & 1.00 & 1.60 & 1.20 & 1.04 \\
    \rowcolor{gray_bg}
    \checkmark &  &  & \checkmark & 1.05 & 1.05 & 1.00 & 1.01 & 1.03 & 1.00 & 1.27 & 1.06 & 1.00 \\

     &  & \checkmark &  & 1.34 & 1.68 & 1.00 & 1.12 & 1.03 & 1.06 & 2.23 & 2.09 & 1.03 \\
     \rowcolor{gray_bg}
     &  & \checkmark & \checkmark & 1.80 & 2.58 & 1.86 & 2.03 & 1.46 & 1.48 & 1.62 & 2.30 & 1.15 \\

    & \checkmark &  &  & 2.49 & 3.47 & 1.96 & 2.67 & 2.71 & 2.08 & 3.40 & 2.03 & 1.29 \\
    \rowcolor{gray_bg}
    & \checkmark &  & \checkmark & 2.57 & 3.32 & 2.11 & 3.07 & 2.87 & 2.03 & 3.49 & 1.88 & 1.32 \\

    \checkmark & \checkmark & \checkmark &  & 2.53 & 3.06 & 1.94 & 2.88 & 2.85 & 1.97 & 3.49 & 2.69 & 1.29 \\
    \rowcolor{gray_bg}
    \checkmark & \checkmark & \checkmark & \checkmark & 2.65 & 3.64 & 2.15 & 3.03 & 3.11 & 1.82 & 3.20 & 2.89 & 1.22 \\

    \bottomrule
    \end{tabular}
    }%
\end{table*}

\mysubsubsec{Effect of data composition.}
As shown in Tab.~\ref{tab:ablation}, I2I-only training fails to generalize to video, while V$\leftrightarrow$I$\leftrightarrow$I transfers image editing priors without direct video supervision, and V2V* provides the strongest spatially-aligned signal. Their contributions are additive: V2V* anchors spatial fidelity while V$\leftrightarrow$I$\leftrightarrow$I broadens coverage of open-domain appearance edits.

\mysubsubsec{Effect of Chain-of-Editing.}
CoE benefits are contingent on chain-structured training. Applied to an I2I-only model, CoE degrades performance as the two-stage format is entirely out of distribution. The V$\leftrightarrow$I$\leftrightarrow$I model benefits most, since its training already includes an intermediate edited frame that CoE exploits as an appearance anchor. Notably, CoE also improves the V2V*-only model (2.49$\to$2.57). Since V2V* is already organized in our decomposed chain format, this shows that CoE generalizes beyond the V$\leftrightarrow$I$\leftrightarrow$I data and serves as a training-free boost for paired video data as well.

\mysubsubsec{Synergy of all components.}
The full model achieves the best overall performance, with CoE gains concentrated in categories where appearance is hardest to specify through text alone, such as global style transfer and creative edits. This confirms CoE functions as a visual grounding mechanism, 
\begin{wrapfigure}{r}{0.5\textwidth}
    \centering
    \vspace{-6pt}
    \includegraphics[width=0.5\textwidth]{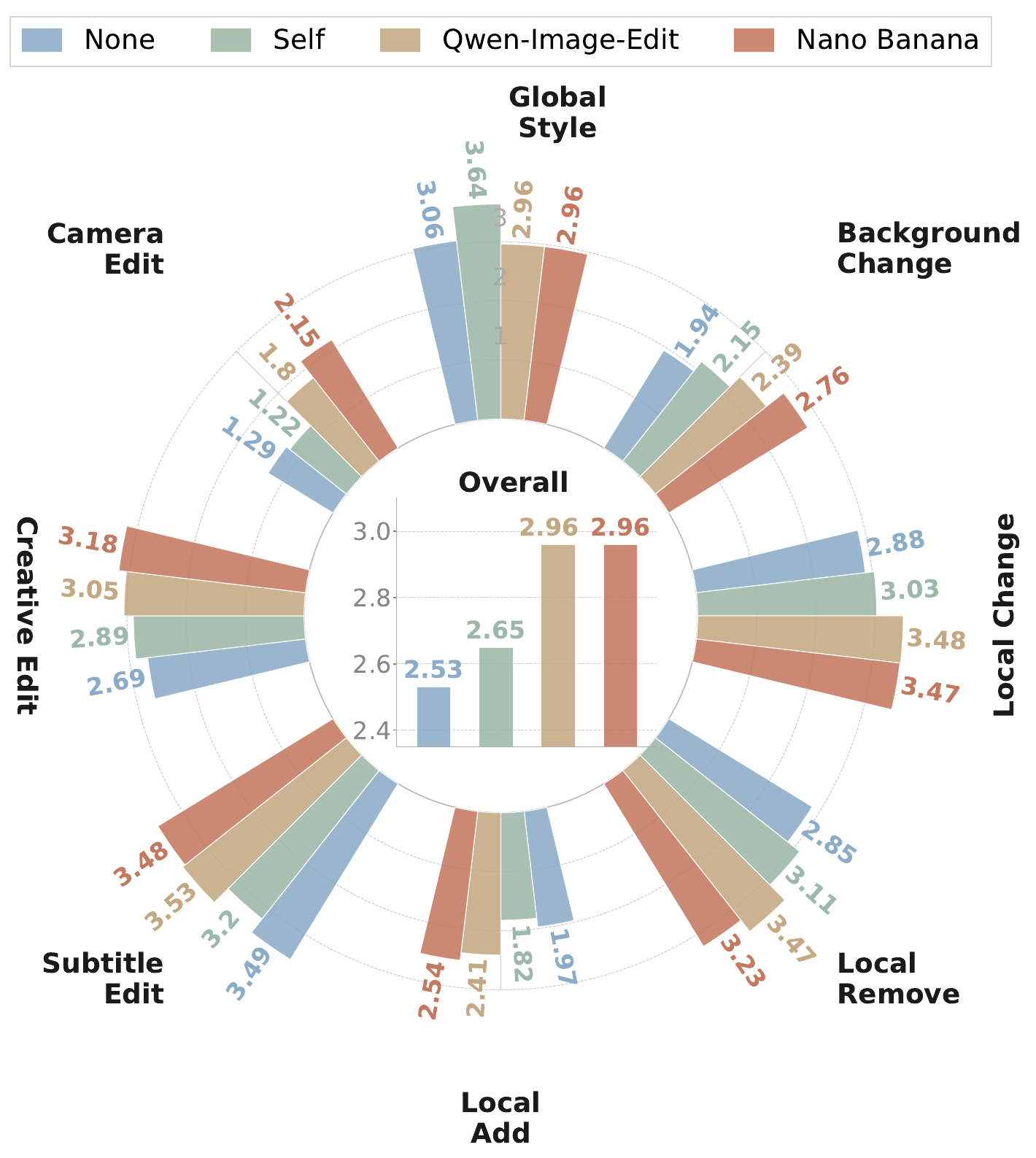}
    \caption{Impact of \textbf{intermediate image editing} quality on CoE video editing performance (OpenVE-Bench, scored by Gemini 2.5 Pro). Stronger external editors (\textit{Qwen-Image-Edit}~\cite{wu2025qwen} and \textit{Nano Banana}~\cite{google2025nanobanana}) consistently outperform self-editing, with complementary strengths highlighting the modular advantage of CoE.}
    \label{fig:abl_coe_qwen_nano}
    \vspace{-8pt}
\end{wrapfigure}
and the three data types together unlock it as an effective strategy without additional training.

\mysubsubsec{Is paired video data the main driver?}
The V2V*-only rows in Tab.~\ref{tab:ablation} already use our decomposed chain format. A \emph{Pure V2V} model (same backbone and V2V data, no decomposition or CoE, in Tab.~\ref{tab:pure_v2v} in Appendix~\ref{sec:app:pure_v2v}) scores only 2.32, below OpenVE-Edit (2.49), which uses a superset\footnote{OpenVE-Edit is trained on the full OpenVE V2V data, whereas our V2V split excludes the Creative Edit and Camera Multi-Shot Edit categories, which are not publicly released.} of our V2V data. In contrast, full data + CoE reaches 2.65 ($+14.2\%$), and $+74.1\%$ on Creative Edit, a category without any direct V2V supervision, while adding only $\sim$6\% visual training tokens. Besides, a blinded pairwise human study on all 431 clips with 10 raters (Fleiss' $\kappa=0.721$) also prefers full data + CoE over V2V*-only + CoE, most clearly Prompt Following (38\% vs.\ 24\%) and Consistency (28\% vs.\ 9\%), showing that the seemingly small MLLM-judged gap between the two (2.57 vs.\ 2.65) is human-perceptible (Appendix~\ref{sec:app:human}).

\mysubsubsec{Effect of intermediate image editing quality.}
Fig.~\ref{fig:abl_coe_qwen_nano} shows how intermediate image quality influences editing performance through the CoE chain.
Self-editing underperforms on categories requiring geometric transformation or novel content synthesis, while replacing it with an external editor yields consistent overall improvement. The two external editors reach the similar overall score with complementary category-level strengths, highlighting a key advantage of the modular I$\to$I interface: advances in image editing transfer directly to video without retraining.

\begin{figure}[t]
    \centering
    \begin{minipage}[t]{0.32\linewidth}
        \centering
        \includegraphics[width=\linewidth]{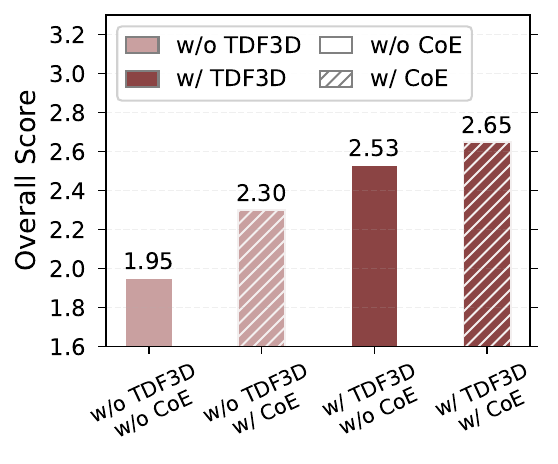}
        \caption{Ablation on TDF3D-RoPE and CoE (Overall score on OpenVE-Bench, scored by Gemini 2.5 Pro). Bar color indicates TDF3D-RoPE; hatch pattern indicates CoE. Both components contribute, and their combination achieves the best result.}
        \label{fig:abl_rope_coe}
    \end{minipage}
    \hfill
    \begin{minipage}[t]{0.32\linewidth}
        \centering
        \includegraphics[width=\linewidth]{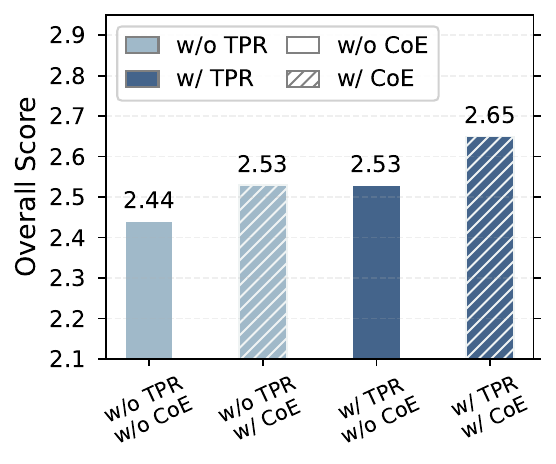}
        \caption{Ablation on TPR and CoE (Overall score on OpenVE-Bench, scored by Gemini 2.5 Pro). Bar color indicates TPR; hatch pattern indicates CoE. Both components contribute, and their combination achieves the best result.}
        \label{fig:abl_tpr_coe}
    \end{minipage}
    \hfill
    \begin{minipage}[t]{0.32\linewidth}
        \centering
        \includegraphics[width=\linewidth]{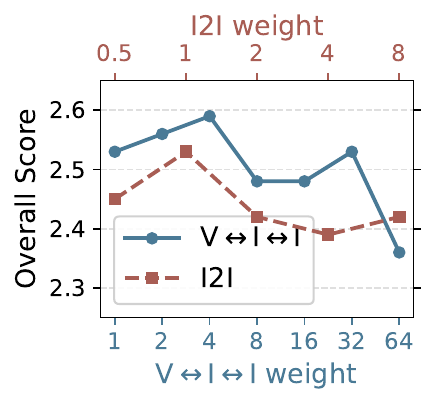}
        \caption{Overall score on OpenVE-Bench scored by Gemini 2.5 Pro as the training data sampling weight of V$\leftrightarrow$I$\leftrightarrow$I (bottom axis, blue) and I2I (top axis, red) are each varied independently, with all other mixture proportions held fixed.}
        \label{fig:mf_weight}
    \end{minipage}
    \vspace{-8pt}
\end{figure}

\mysubsubsec{Effect of TDF3D-RoPE.}
Fig.~\ref{fig:abl_rope_coe} summarizes the contributions of TDF3D-RoPE and CoE.
Both components improve overall performance, and their combination achieves the best score of 2.65.
TDF3D-RoPE provides the larger standalone gain by establishing pixel-level spatial correspondence across shots, a prerequisite for faithful localized edits.
CoE further refines edits by routing generation through an explicit intermediate image, and the two components together are complementary.
Detailed per-category analysis is provided in Appendix~\ref{sec:app:additional_exp}.

\mysubsubsec{Effect of Temporal Position Randomization.}
Fig.~\ref{fig:abl_tpr_coe} shows that TPR and CoE interact multiplicatively. Without TPR, a fixed temporal index creates a spurious structural correlation that prevents CoE's intermediate keyframe from serving as a genuine appearance anchor. TPR eliminates this bias, and only their combination reaches peak performance. Per-category details are in Appendix~\ref{sec:app:additional_exp}.

\mysubsubsec{Effect of training data sampling weights.}
Fig.~\ref{fig:mf_weight} shows that each data source contributes complementary supervision, and performance is best when they are balanced. Moderately upsampling V$\leftrightarrow$I$\leftrightarrow$I improves chain exposure, while overly large weights reduce paired video supervision. I2I performs best at the default ratio. Overall, the default mixture provides a good trade-off among the three sources.

\subsection{Qualitative Results}

\begin{figure*}[t]
    \centering
    \includegraphics[width=\textwidth]{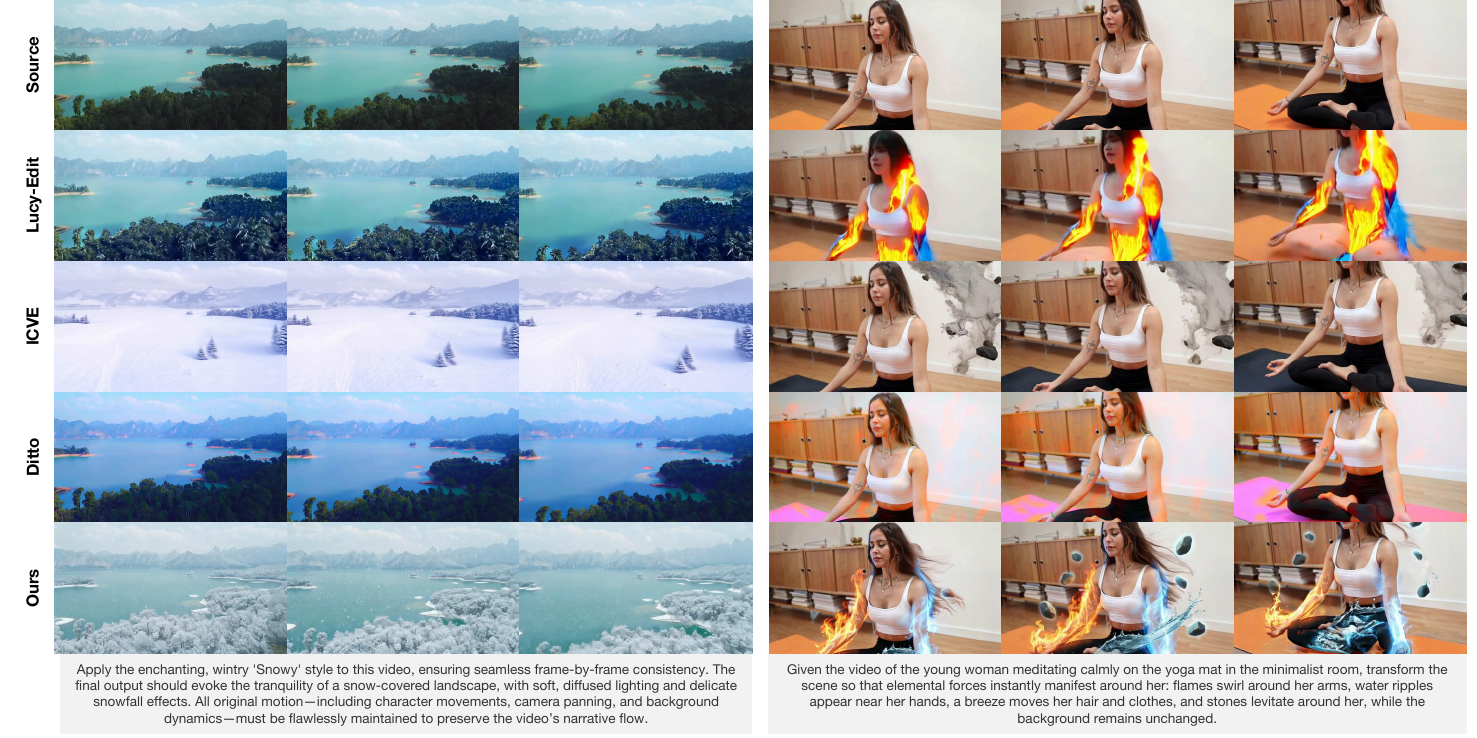}
    \caption{Qualitative results on global scene stylization (\textbf{left}, summer-to-snow) and localized compositional editing (\textbf{right}, injecting fire/water/wind). }
    \label{fig:case_study}
\end{figure*}

\begin{figure*}[t]
    \centering
    \includegraphics[width=\textwidth]{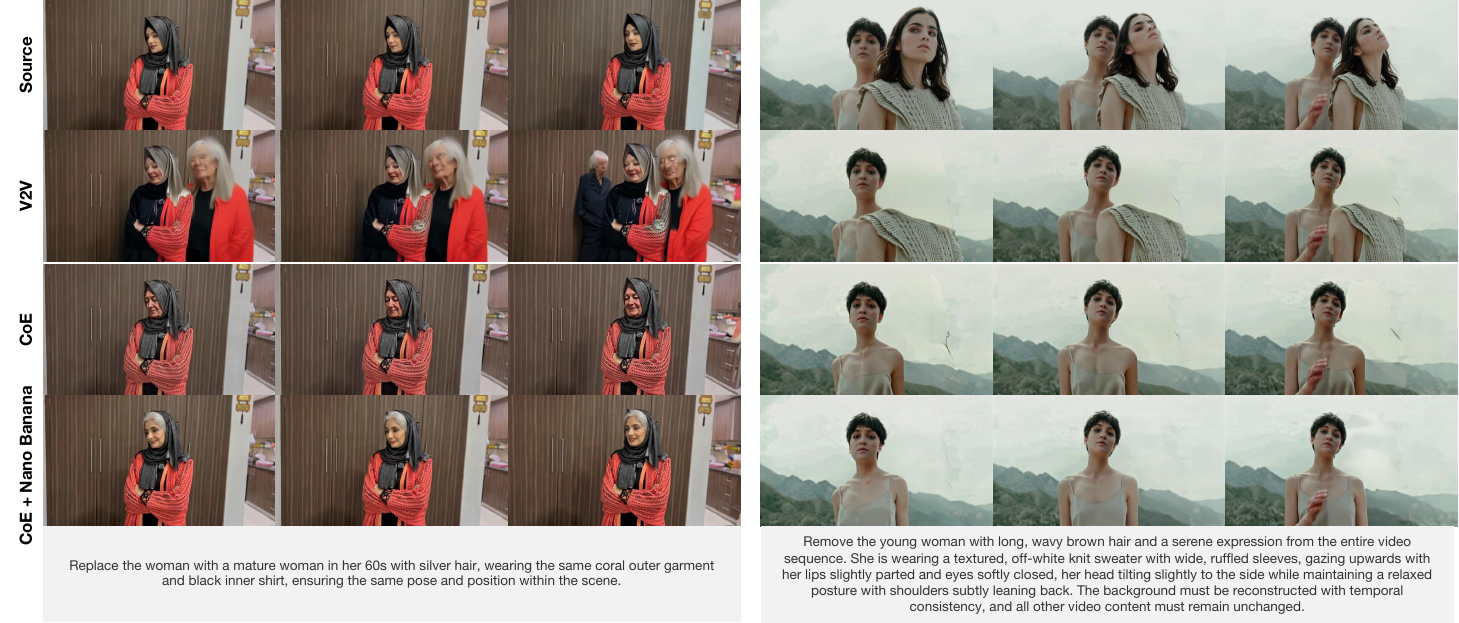}
    \caption{Comparison of V2V direct inference, CoE with self editing, and CoE with an external editor on subject replacement (\textbf{left}) and removal (\textbf{right}). }
    \label{fig:case_study_ablation}
\end{figure*}

Fig.~\ref{fig:case_study} shows global scene stylization (summer-to-snow) and localized compositional effects (fire, water, wind), both maintaining temporal consistency without task-specific fine-tuning.
Fig.~\ref{fig:case_study_ablation} isolates the CoE contribution on subject replacement and removal.
V2V direct inference produces flickering and ghosting, while CoE with self-editing eliminates this instability by anchoring to a reference image.
Routing through an external editor further improves semantic accuracy with no retraining, demonstrating that the modular I$\to$I interface lets image editing advances transfer directly to video.

\section{Conclusion}

In this paper, we presented \ours, a video editing framework that substantially reduces the dependence on large-scale paired video data by routing editing intent through the image domain via a structured $\text{V}{\to}\text{I}{\to}\text{I}{\to}\text{V}$ chain, enabling scalable joint training from heterogeneous image and video corpora under a unified flow-matching objective.
TDF3D-RoPE propagates pixel-level appearance changes faithfully across every frame, while Chain-of-Editing provides a principled test-time scaling mechanism that trades compute for fidelity without retraining.
Evaluations on OpenVE-Bench confirm state-of-the-art performance among state-of-the-art methods, and we hope this progressive editing paradigm, \ie, grounding video editing in mature image priors, opens a practical and scalable path for future complex video creation research.

\begin{ack}
This research is supported by the National Research Foundation, Singapore under its National Large Language Models Funding Initiative (AISG Award No: AISG-NMLP-2024-002). Any opinions, findings and conclusions or recommendations expressed in this material are those of the author(s) and do not reflect the views of National Research Foundation, Singapore.
\end{ack}

\bibliographystyle{plainnat}
\bibliography{references}

\appendix

\newpage

\addcontentsline{toc}{section}{Appendix}

\renewcommand{\contentsname}{Contents}

\begingroup
\setcounter{tocdepth}{2}
\tableofcontents
\endgroup

\clearpage

\section{Additional Details of Methodology}\label{sec:app:method}

\subsection{TDF3D-RoPE: Position Assignment Formulas}

We partition the interleaved sequence into \emph{shots}, each consisting of one text segment followed by one visual segment.
A global offset $\tau_{\mathrm{start}}$ is accumulated across shots.
Within shot $k$ of text length $l_k$, text tokens receive sequential 1D positions $[\tau_{\mathrm{start}},\, \tau_{\mathrm{start}} + l_k)$, broadcast to all three axes $(p_t, p_h, p_w)$.
For the visual segment, floating-point frame timestamps $\{t_i\} \subset \mathbb{R}_{\geq 0}$ serve as temporal positions, natively accommodating non-uniform sampling and variable frame rates.
All three axes of the visual tokens are uniformly shifted by $\tau_{\mathrm{start}} + l_k$, keeping every shot in a non-overlapping position range.
The offset advances as:
\begin{equation}
    \tau_{\mathrm{start}} \leftarrow \tau_{\mathrm{start}} + l_k + \max\!\bigl(\lceil t_{\mathrm{last}} \rceil,\, H,\, W\bigr),
\end{equation}
where $t_{\mathrm{last}}$ is the last frame timestamp of the current shot, and $H$, $W$ are the spatial grid dimensions.
These assignments constitute the \textbf{inter-shot component}.

For the \textbf{intra-video component}, visual tokens are assigned their raw positions $(t_i, h, w)$ with zero global offset, independent of shot index; text tokens are fixed at position $-1$.
Two patches at the same spatial location $(h, w)$ in any pair of shots therefore share an identical intra-video $(p_h, p_w)$ and have zero relative spatial offset, making their attention scores purely content-driven.
The head dimension $d$ is partitioned equally across the three axes ($d/3$ channels each), and the final per-token frequency is the element-wise sum of the inter-shot and intra-video components.

\subsection{Chain-of-Editing: Implementation Details}

\mysubsubsec{Stage 1 — Image editing (V$\to$I$\to$I).}
The interleaved context is assembled from three shots: the source video $\mathbf{V}^s$, the source keyframe $\mathbf{I}^s$, and the target image slot to be denoised.
Frame indices are assigned using the actual temporal timestamps of the selected keyframes, normalized by the total video length.
The visual segments are processed frame-independently (\texttt{frame\_separate=True}), reflecting the spatial nature of image editing.

\mysubsubsec{Stage 2 — Video generation (I$\to$V).}
The interleaved context contains four shots: $\mathbf{V}^s$, $\mathbf{I}^s$, and $\hat{z}^t$ from Stage~1 (three conditioning shots), followed by the target video slot.
The four-shot frame index sequence is $\bigl[\,\{0,\ldots,T{-}1\},\;\{t_i^{\mathrm{mf}}\},\;\{t_i^{\mathrm{mf}}\},\;\{0,\ldots,T{-}1\}\,\bigr]$, aligning the target video temporally with the source.
Visual segments are processed jointly as a continuous sequence (\texttt{frame\_separate=False}) to model cross-frame dependencies.

\mysubsubsec{Per-shot timestep conditioning.}
This design keeps the architecture identical to training, where reference segments are likewise observed as clean tokens in the interleaved sequence.
No architectural modifications—such as additional cross-attention modules or conditioning adapters—are required at inference time.

\mysubsubsec{Test-time scaling.}
Longer chains—\eg, iterating the image-editing stage multiple times before advancing to video generation—progressively enrich the interleaved context and allow the model to refine intermediate results, trading additional compute for higher editing fidelity.
Because the chain extension operates purely by prepending more conditioning shots to the context, it requires no retraining or fine-tuning.

\section{In-context Video-based Image Editing Annotation}\label{app:annotation}
This section describes the pipeline for annotating and constructing the V$\leftrightarrow$I$\leftrightarrow$I (video-to-image-to-image) training data, where each sample pairs an original video frame with an edited counterpart under a text instruction.
The annotation process involves three coordinated stages, each with explicit design constraints to ensure data quality.
First, \textbf{category annotation} is performed before instruction writing to control the distribution of editing types: a live feedback mechanism steers GPT-4o toward under-represented categories, preventing the long-tailed imbalance that would arise from unconstrained annotation.
Second, \textbf{category selection rules} are applied so that the assigned editing type is grounded in what is actually visible in the raw video, ensuring the suggested edit aligns with natural human aesthetics and realistic editing intent rather than being arbitrary or implausible.
Third, \textbf{text instruction constraints} are imposed during annotation to encourage lexical and semantic diversity: by tracking annotation statistics and injecting them into the prompt, we discourage the model from repeatedly gravitating toward salient but overused edits (\eg, always annotating sunset or color-grading effects), producing a richer and more varied instruction vocabulary across the dataset.

\subsection{Editing Category Analysis}\label{app:category_ann}

To ensure a balanced and diverse editing category distribution across the dataset, we run a dedicated category analysis pass on each raw video before the full instruction-annotation stage.
Given $N\!=\!5$ frames uniformly sampled from a video, \textbf{GPT-4o} is prompted to (i)~produce a structured multi-granularity caption covering the overall scene, individual characters, prominent objects, and camera behavior, and (ii)~propose 1--5 applicable editing categories with subcategories and motivations drawn from a predefined 13-category taxonomy: \texttt{remove}, \texttt{replace}, \texttt{attribute}, \texttt{background}, \texttt{global}, \texttt{effect}, \texttt{camera}, \texttt{action}, \texttt{expression}, \texttt{dynamics}, \texttt{position}, \texttt{pose}, and \texttt{orientation}.

A key design feature is a \textbf{live distribution feedback} mechanism.
As annotations accumulate, a process-safe shared dictionary tracks the running per-category count.
At each inference call, the prompt is dynamically populated with three derived statistics: (a)~the current category distribution (\texttt{category\_dist}), (b)~the five most over-represented categories (\texttt{high\_category}), and (c)~the five most under-represented categories (\texttt{low\_category}).
Injecting this live context steers the model toward less-covered categories, progressively balancing the taxonomy without any manual curation or post-hoc resampling.

The category labels produced in this step are used downstream to guide the text instruction annotation stage (Appendix~\ref{app:text_ann}), where each sample is constrained to one editing category and the same low-frequency steering statistics are re-injected to maintain global balance.

\begin{tcolorbox}[colback=gray!5,colframe=gray!90,title=Prompt for Editing Category Analysis,breakable]
{\footnotesize
You are a professional video editing assistant.
Given \{N\_FRAME\} frames from a video, your tasks are:\\
1.~Describe the video content: overall scene, characters (appearance, relations), objects, and camera.\\
2.~List 1--5 applicable editing categories with subcategory and motivation, chosen from:
\texttt{remove}, \texttt{replace}, \texttt{attribute}, \texttt{background}, \texttt{global}, \texttt{effect}, \texttt{camera}, \texttt{action}, \texttt{expression}, \texttt{dynamics}, \texttt{position}, \texttt{pose}, \texttt{orientation}.\\[2pt]
\textbf{Selection rules} (prefer categories supported by visible evidence):\\
\textit{Humans \& faces}\\
$\bullet$~Large/clear face (close-up) $\to$ \texttt{expression} (smile, surprise, anger), then \texttt{orientation} (look direction), then \texttt{effect} (tears, sweat, glow).\\
$\bullet$~Upper body visible $\to$ \texttt{action} (wave, point, clap) and interaction (handing/holding), then \texttt{pose}.\\
$\bullet$~Full body visible $\to$ \texttt{pose} (stance, posture), then \texttt{action}, then \texttt{position} (move left/right).\\
$\bullet$~Multiple people $\to$ \texttt{action}/interaction (hug, handshake), then \texttt{position} (swap places), then \texttt{expression} (match reactions).\\
$\bullet$~Occluded/crowded $\to$ \texttt{remove}/\texttt{replace} character only if practical; otherwise \texttt{camera} (reframe) or \texttt{background} (blur crowd).\\[2pt]
\textit{Objects \& props}\\
$\bullet$~Single prominent object $\to$ \texttt{remove}/\texttt{replace} object, then \texttt{attribute} (color, material, texture), then \texttt{effect} (transform).\\
$\bullet$~Many repeated objects $\to$ \texttt{position} (count/layout), then \texttt{remove} clutter.\\
$\bullet$~Text/logos/signage visible $\to$ \texttt{remove}/\texttt{replace} text; fall back to \texttt{camera} (crop/zoom).\\
$\bullet$~Unsafe/sensitive items $\to$ \texttt{remove}/\texttt{replace} object (make non-threatening).\\[2pt]
\textit{Scene \& background}\\
$\bullet$~Distracting or iconic background $\to$ \texttt{background} (replace), or \texttt{remove} distracting objects, then \texttt{camera} (reframe).\\
$\bullet$~Strong location context (beach, street) $\to$ \texttt{global} (weather/lighting/season/style), only if consistent with visible cues (shadows, clothing).\\
$\bullet$~Indoor with lamps/windows $\to$ \texttt{global} (lighting direction, contrast).\\[2pt]
\textit{Motion \& dynamics}\\
$\bullet$~Natural phenomena (fire, smoke, water, rain, fog) $\to$ \texttt{dynamics} (intensify/slow/stop), then \texttt{global} (stormy/clear), then \texttt{effect} (sparks, embers).\\
$\bullet$~Vehicles / fast motion $\to$ \texttt{camera} (tracking, stabilization) and \texttt{dynamics} (skid, splash, dust trail).\\
$\bullet$~Explosions / breaking / spills / collapsing $\to$ \texttt{dynamics} first, then \texttt{effect}.\\[2pt]
\textit{Camera \& style}\\
$\bullet$~Shaky footage $\to$ \texttt{camera} (stabilize, smooth transitions).\\
$\bullet$~Subject too small / too wide $\to$ \texttt{camera} (crop-in, push-in, reframe).\\
$\bullet$~Rapid cuts / inconsistent angles $\to$ \texttt{camera} (match cuts), then \texttt{global} (color consistency).\\[2pt]
\textit{Global mood}\\
$\bullet$~Night/dark cues $\to$ \texttt{global} (darker atmosphere, stronger contrast); bright daylight $\to$ \texttt{global} (clearer, punchier look).\\
$\bullet$~Prefer concrete global edits: lighting direction, fog, rain, storm, season, cinematic style. Avoid contradicting clothing or shadow cues.\\[2pt]
\textit{Effects}\\
$\bullet$~Choose \texttt{effect} for a localized, visually obvious transformation (glow, freeze, dissolve, morph) on one focal element (face, hand, single object).\\[2pt]
\textbf{Balance constraint}: current category distribution -- \{category\_dist\}. Avoid over-represented categories; prefer under-represented ones: \{low\_category\}.\\[2pt]
\textbf{Output}: a single JSON object -- \{``caption'': \{``video'':~$\langle\cdot\rangle$, ``scene'':~$\langle\cdot\rangle$, ``char1'':~$\langle\cdot\rangle$, \ldots\}, ``editing1'': \{``category'':~$\langle\cdot\rangle$, ``subcategory'':~$\langle\cdot\rangle$, ``motivation'':~$\langle\cdot\rangle$\}, ``editing2'': \{\ldots\}, \ldots\}
}
\end{tcolorbox}

\subsection{Text Instruction Annotation}\label{app:text_ann}

Each video clip in the dataset is pre-assigned an editing category from the 13-category taxonomy described in Appendix~\ref{app:category_ann}.
To generate text editing instructions grounded in actual video content, we uniformly sample $N$ frames from each clip and send them to \textbf{GPT-4o} together with a category-specific prompt.

The model is instructed to: (1)~carefully describe the visual content of the sampled frames, including the scene, characters, objects, and camera; (2)~propose a plausible and visually meaningful editing operation consistent with the assigned category; (3)~formulate a concise forward editing instruction; and (4)~formulate a \emph{reversed} instruction that describes how to edit the result back toward the original, \emph{without} assuming any prior knowledge of the original video state (\eg, if the forward instruction is ``remove the dog,'' the reversed instruction is ``add a dog'').
The model additionally identifies which frames are affected by the edit.

The output is a structured JSON record containing the video caption, editing motivation, forward instruction, reversed instruction, and the set of affected frame indices.
The reversed instructions are included to support bidirectional training, ensuring the model learns both how to apply an edit and how to undo it without relying on privileged knowledge of the source content.

The 13 editing categories each have a distinct scope and constraint, summarized as follows (these definitions are used as \textit{Definition of \{CATEGORY\}} in the prompt):
\begin{itemize}
  \item \textbf{Remove}: completely eliminate a visible character, animal, or object so it is absent from every frame in which it originally appeared.
  \item \textbf{Replace}: substitute an existing subject with a different one while preserving the original motion, timing, camera movement, and scene layout.
  \item \textbf{Attribute}: modify intrinsic properties of an existing object (color, texture, material, brightness, or shape) without adding, removing, or replacing the object itself; the change is applied consistently across all frames.
  \item \textbf{Background}: swap the environment behind the main foreground subject(s) throughout the entire clip, leaving foreground content untouched.
  \item \textbf{Global}: apply a uniform visual transformation to the whole video (such as color grading, lighting style, season, weather, or artistic look), affecting every frame equally without targeting any specific element.
  \item \textbf{Effect}: overlay a localized visual treatment (glow, particle, distortion, hologram, motion trail, \emph{etc.}) on a specific subject or object, sustained consistently across frames.
  \item \textbf{Camera}: simulate post-production virtual camera movements (zoom, pan, dolly, tilt, rotation, or handheld shake) that alter the viewer's perspective without modifying scene content.
  \item \textbf{Action}: change what the main subject is \emph{doing} (modifying motion type (\eg, walk $\to$ run), adding continuous movement to a static subject, or altering motion intensity) while leaving appearance and scene unchanged.
  \item \textbf{Expression}: alter the visible emotional state of a character's face (\eg, neutral $\to$ smile, calm $\to$ angry) through adjustments to brow, eye, and mouth configuration, without changing action or scene.
  \item \textbf{Dynamics}: introduce a physical state change in an object over time (such as igniting, extinguishing, melting, blooming, exploding, or dissolving), emphasizing the transformation process itself.
  \item \textbf{Position}: reposition a subject within the frame (left/right/up/down) to alter composition, without changing its size, color, or temporal behavior.
  \item \textbf{Pose}: reconfigure the body arrangement of a character (limb placement, stance, or body posture) without changing facial expression, spatial orientation, or camera angle.
  \item \textbf{Orientation}: rotate or flip a subject to change the direction it faces or the angle at which it appears, without altering pose, expression, or scene content.
\end{itemize}

\begin{tcolorbox}[colback=gray!5,colframe=gray!90,title=Prompt for Text Instruction Annotation,breakable]
{\footnotesize
You are a professional video editing assistant.
You will be provided with \{N\_FRAME\} frames from a video and an assigned editing type: \textbf{\{CATEGORY\}}.\\[2pt]
\textit{Definition of \{CATEGORY\}}: \{CATEGORY\_DEFINITION\}\\[2pt]
Your task is to write a text instruction to edit the video following the assigned editing type:\\
1.~Carefully observe the frames and describe the content: environment, main subjects (appearance, relations, interactions), and background. Identify characters and objects without using real names.\\
2.~Following the editing type, propose how to edit the video to make it more engaging, explaining your reasoning.\\
3.~Summarize the editing idea into a concise user instruction.\\
4.~Write a reversed instruction describing how to edit the result back, \emph{without} any knowledge of the original state (\eg, ``remove the dog'' $\to$ reversed: ``add a dog''). Do \textbf{not} use words like ``restore'', ``revert'', or ``return''.\\
5.~Identify which frames should be edited (\eg, only frames where the subject appears).\\[2pt]
\textbf{Note}: Each frame must be individually editable; the instruction must be valid when applied per frame.\\[2pt]
\textbf{Output}: a single JSON object in one line --\\
\{``caption'': \{``video'':~$\langle$description$\rangle$, ``scene'':~$\langle\cdot\rangle$, ``char1'':~$\langle\cdot\rangle$, ``obj1'':~$\langle\cdot\rangle$, ``camera'':~$\langle\cdot\rangle$\}, ``motivation'':~$\langle$what and why to edit$\rangle$, ``instruction'':~$\langle$concise editing instruction$\rangle$, ``instruction\_reversed'':~$\langle$reversed instruction$\rangle$, ``frame\_idx\_edited'':~$\langle$list of frame indices $<$\{N\_FRAME\}$\rangle$\}
}
\end{tcolorbox}

\subsection{Frame Editing Annotation}\label{app:frame_editing_ann}

Given the text editing instruction produced in Appendix~\ref{app:text_ann}
and the corresponding set of affected frame indices,
we apply \textbf{Seedream~4.5} as an instruction-guided image editor
to produce an edited counterpart for each selected keyframe.
Each frame is edited independently: the text instruction is passed to
Seedream~4.5 together with a concise constraint that the model should
maintain the original composition and background,
preserve the main subject's identity,
and produce edits that are subtle yet clearly visible.
This per-frame formulation is consistent with the design of the
text annotation stage (Appendix~\ref{app:text_ann}),
where each instruction is required to be self-contained and
valid when applied to a single image.
The final output for each clip is a set of
\emph{(original frame, edited frame, instruction)} triplets
over the affected indices, which serve as the video-based in-context image editing data for training our model. 
\section{Detailed Experimental Settings}\label{sec:app:exp_setting}

\subsection{Training Data}

\ours is trained on three complementary data sources, which correspond to the \textbf{I2I}, \textbf{V2V*}, and \textbf{V$\leftrightarrow$I$\leftrightarrow$I} splits used in the ablation study (Tab.~\ref{tab:ablation}).
Their sizes are 1.20M image editing pairs (I2I, OmniEdit~\cite{wei2024omniedit}), 2.45M video editing pairs (V2V, OpenVE~\cite{openve3m2024}), and 1.63M chain triplets (V$\leftrightarrow$I$\leftrightarrow$I, ours). Our pairwise video split is a subset of the training set of OpenVE-Edit~\cite{openve3m2024}, as it excludes the unreleased Creative Edit and Camera Multi-Shot Edit categories.

\mysubsubsec{Image-to-Image data (I2I).}
\textbf{I2I} uses paired image editing data from OmniEdit~\cite{wei2024omniedit}, covering a diverse range of editing categories with high-quality before–after image pairs.

\mysubsubsec{Video-to-Video data (V2V*).}
\textbf{V2V*} uses paired video editing data from OpenVE~\cite{openve3m2024}, where each sample consists of a source video, an editing instruction, and a corresponding edited video.
The asterisk indicates that these pairs are not used as plain video-to-video samples: during training, each pair is reformatted into a randomly chosen sub-chain of $\text{V}\!\to\!\text{I}\!\to\!\text{I}\!\to\!\text{V}$, including direct video-to-video editing, multi-frame-to-multi-frame editing (frames sampled from both videos), multi-frame-to-video generation, and the full interleaved chain.
This distinguishes V2V* from the Pure V2V baseline (Appendix~\ref{sec:app:pure_v2v}), which trains on the same pairs without decomposition.

\mysubsubsec{Video$\leftrightarrow$Image$\leftrightarrow$Image chain data (V$\leftrightarrow$I$\leftrightarrow$I).}
This is the core novel data source that enables \ours to bridge video editing through the image domain.
We start from raw videos in the OpenVE collection \emph{without} using any of their pairwise editing annotations.
For each video, we uniformly sample $N\!=\!5$ frames and feed them to GPT-4o with a structured prompt that generates a balanced editing instruction across 12 categories (\eg, global style, local add/remove, background, dynamics).
The instruction is then applied independently to each sampled frame using Seedream~4.5~\cite{seedream2025}, an expressive image editor, to produce per-frame before–after pairs.
Together with the original video, these form a $\text{Video}\!\to\!\text{Image}\!\to\!\text{Image}$ chain triplet: the source video, an unedited representative frame, and its edited counterpart.
This construction avoids the need for any paired video editing supervision while providing rich, instruction-grounded visual demonstrations for training.

\subsection{Benchmark Details}

\mysubsubsec{Benchmark.}
We evaluate on OpenVE-Bench~\cite{openve3m2024}, a comprehensive benchmark for instruction-guided video editing comprising 431 video clips organized into eight editing categories.
Six categories involve \emph{spatially-aligned} edits, where the output must preserve the original spatial layout: \textbf{Global Style} (transferring artistic or weather styles globally), \textbf{Background Change} (replacing scene backgrounds while keeping foreground subjects intact), \textbf{Local Change} (modifying the appearance of a specific object or region), \textbf{Local Remove} (erasing a target element and plausibly filling the vacated region), \textbf{Local Add} (inserting a new object into the scene), and \textbf{Subtitle Edit} (modifying on-screen text overlays).
The remaining two categories involve \emph{non-spatially-aligned} edits that do not require strict spatial correspondence: \textbf{Creative Edit} (free-form stylistic or compositional transformations) and \textbf{Camera Edit} (synthesizing novel viewpoints or camera motions).
This taxonomy spans a wide spectrum of real-world editing demands, from fine-grained local manipulations to global cinematic effects.

\mysubsubsec{Evaluation protocol.}
Following OpenVE-Bench, we adopt an MLLM-based evaluation protocol that closely mirrors human judgment: the MLLM is prompted with the same rubric and scoring dimensions a human rater would use. All benchmark scores reported in our tables are produced automatically by the MLLM judge; the separate human study is described in Appendix~\ref{sec:app:human}.
For each test sample, the source video, editing instruction, and edited video are jointly fed to a multimodal evaluator, which scores the output along three orthogonal dimensions on a \textbf{1--5} scale:
(i)~\textit{Instruction Compliance}: how faithfully the edit follows the specified instruction;
(ii)~\textit{Consistency \& Detail Fidelity}: how well unedited regions and fine-grained visual details are preserved; and
(iii)~\textit{Visual Quality \& Stability}: overall perceptual quality and temporal coherence.
Crucially, Instruction Compliance acts as an upper bound for the remaining two dimensions: a visually polished but semantically incorrect edit is penalized even if it scores high on quality alone.
Per-category scores are averaged across all three dimensions to yield a final overall score, and we report both category-level and overall performance.

\mysubsubsec{Baselines.}
We compare \ours against a range of representative methods covering both open-source and commercial systems.
Open-source baselines include VACE~\cite{jiang2025vace}, a unified video creation and editing framework; OmniVideo~\cite{omnivideo2024}, an omni-directional video editing model; InsViE~\cite{insvie2024}, an instruction-driven video editor; Lucy-Edit~\cite{lucyedit2024}, a training-efficient video editing method; ICVE~\cite{icve2024}, an in-context video editor; DITTO~\cite{ditto2024}, a disentangled video editing approach; and OpenVE-Edit~\cite{openve3m2024}, the official model released alongside OpenVE-Bench.
We also compare against the proprietary system \textbf{Runway Aleph}, which represents the current commercial frontier.
All methods are evaluated under their default inference settings on the full OpenVE-Bench test set.

\mysubsubsec{Baseline selection.}
Our selection follows three criteria.
(1)~\emph{Family coverage}: we pick one or two representative methods per active family of instruction-guided video editing (unified creation/editing frameworks, in-context editors, and dataset-centric editors) rather than several methods from the same family.
(2)~\emph{Recency}: all baselines were released within roughly one year before our submission, \ie, VACE (Mar.~2025), InsViE (Mar.~2025), OmniVideo (Jul.~2025), Lucy-Edit (Sep.~2025), ICVE (Oct.~2025), DITTO (Oct.~2025), and OpenVE-Edit (Dec.~2025).
(3)~\emph{Backbone fairness}: these baselines build on modern DiT-based video foundation models (\eg, VACE and Lucy-Edit on Wan~\cite{wan2025}, ICVE on HunyuanVideo~\cite{kong2024hunyuan}), so the comparison isolates our technical contribution rather than gains from a newer backbone.
AnyV2V~\cite{ku2024anyv2v} (Mar.~2024) is built on the U-Net-based I2VGen-XL~\cite{zhang2023i2vgenxl}, but since it is the most closely related image-editing-driven method, we additionally compare against it in Appendix~\ref{sec:app:anyv2v}.
EditVerse~\cite{li2024editverse} is not open-sourced and therefore cannot be evaluated.

\subsection{Implementation Details}\label{sec:app:impl}

Tab.~\ref{tab:impl} consolidates the architecture, data, training, and inference settings of \ours. Training and inference launch scripts and full hyperparameter configurations are provided in our code repository.

\begin{table}[ht]
    \centering
    \small
    \caption{Implementation details of \ours.}
    \label{tab:impl}
    \begin{tabular}{ll}
    \toprule
    \multicolumn{2}{l}{\textit{Architecture}} \\
    DiT backbone & 3B MM-DiT, initialized from the same T2V-pretrained backbone as VINCIE~\cite{qu2025vincie} \\
    VAE & Video VAE inflated from SD3~\cite{esser2024sd3}; spatial $\times 8$, temporal $\times 4$, 16 channels \\
    Patch size $(t, h, w)$ & $(1, 2, 2)$ \\
    Text encoder & Flan-T5~\cite{chung2024flant5} \\
    Trainable / frozen & DiT trained; VAE and text encoder frozen \\
    \midrule
    \multicolumn{2}{l}{\textit{Training data}} \\
    V2V (OpenVE) & 2.45M pairs \\
    I2I (OmniEdit) & 1.20M pairs \\
    V$\leftrightarrow$I$\leftrightarrow$I (ours) & 1.63M triplets \\
    \midrule
    \multicolumn{2}{l}{\textit{Training}} \\
    Hardware & 32 $\times$ H100 GPUs \\
    Total training & $\approx$150 hours, 42k steps \\
    Stage 1 & $256\times256$, $\approx$80 hours \\
    Stage 2 & $480\times640$, $\approx$70 hours, initialized from Stage 1 \\
    Learning rate & $5\times10^{-5}$ \\
    \midrule
    \multicolumn{2}{l}{\textit{Inference}} \\
    Output frames & 65 (maximum multi-shot budget) \\
    FPS & 12 \\
    Sampling steps & 32 \\
    CFG scale & 2.5 \\
    Source keyframe $\mathbf{I}^s$ & Temporal mid-frame of $\mathbf{V}^s$ \\
    \bottomrule
    \end{tabular}
\end{table}

\section{Additional Experimental Results}\label{sec:app:additional_exp}

\subsection{Performance Comparison Evaluated by Seed1.6-VL}\label{sec:app:seedvl}

Tab.~\ref{tab:perf_seedvl} reports full per-category results evaluated by Seed1.6-VL as an independent MLLM judge.
The rank ordering is consistent with the Gemini 2.5 Pro results in the main paper: \ours achieves an overall score of \textbf{3.09}, surpassing all open-source baselines by a substantial margin and narrowing the gap to Runway Aleph to $0.41$ points.
Category-level trends are likewise stable: Global Style (4.25) and Creative Edit (3.39) are consistent strengths, while Camera Edit (2.01) remains the primary weakness, supporting the reliability of our main conclusions.

\begin{table*}[ht]
    \centering
    \small
    \setlength{\tabcolsep}{3pt}
    \caption{Performance comparison of video editing methods across eight editing categories, evaluated by \textbf{Seed1.6-VL}. Scores are automatic MLLM ratings on a 1--5 scale following the OpenVE-Bench protocol. \#Reso. denotes the output resolution. \color{mygray}{Grey} rows denote closed-source commercial models.}
    \label{tab:perf_seedvl}
    \resizebox{0.95\textwidth}{!}{%
    \begin{tabular}{l|c|c|cccccccc}
    \toprule
    Methods & \#Reso. & Overall & \begin{tabular}[c]{@{}c@{}}Global\\ Style\end{tabular} & \begin{tabular}[c]{@{}c@{}}Background\\ Change\end{tabular} & \begin{tabular}[c]{@{}c@{}}Local\\ Change\end{tabular} & \begin{tabular}[c]{@{}c@{}}Local\\ Remove\end{tabular} & \begin{tabular}[c]{@{}c@{}}Local\\ Add\end{tabular} & \begin{tabular}[c]{@{}c@{}}Subtitle\\ Edit\end{tabular} & \begin{tabular}[c]{@{}c@{}}Creative\\ Edit\end{tabular} & \begin{tabular}[c]{@{}c@{}}Camera\\ Edit\end{tabular} \\
    \midrule
    \color{mygray}{Runway Aleph} & \color{mygray}{$1280 \times 720$} & \color{mygray}{3.50} & \color{mygray}{3.47} & \color{mygray}{2.84} & \color{mygray}{3.88} & \color{mygray}{3.88} & \color{mygray}{2.79} & \color{mygray}{3.50} & \color{mygray}{3.23} & \color{mygray}{4.48} \\
    VACE~\cite{jiang2025vace} & $1280 \times 720$ & 1.17 & 1.41 & 1.16 & 1.43 & 1.00 & 1.05 & 1.02 & 1.13 & 1.16 \\
    OmniVideo~\cite{omnivideo2024} & $640 \times 352$ & 1.02 & 1.02 & 1.00 & 1.00 & 1.00 & 1.00 & 1.16 & 1.00 & 1.00 \\
    InsViE~\cite{insvie2024} & $720 \times 480$ & 1.40 & 2.25 & 1.23 & 1.60 & 1.00 & 1.23 & 1.22 & 1.68 & 1.02 \\
    Lucy-Edit~\cite{lucyedit2024} & $1280 \times 704$ & 1.95 & 2.17 & 2.20 & \underline{3.30} & 1.03 & \underline{2.37} & 1.06 & 2.35 & 1.14 \\
    ICVE~\cite{icve2024} & $384 \times 240$ & 2.25 & 2.35 & 1.86 & 2.91 & \underline{2.68} & 2.27 & 2.04 & 1.94 & 1.38 \\
    DITTO~\cite{ditto2024} & $832 \times 480$ & 2.06 & \underline{3.70} & 2.23 & 2.28 & 1.00 & 2.08 & 1.01 & \underline{2.61} & \underline{1.51} \\
    OpenVE-Edit~\cite{openve3m2024} & $1280 \times 704$ & \underline{2.41} & 3.11 & \textbf{2.72} & 3.19 & 1.42 & \textbf{2.41} & \underline{2.56} & 2.01 & 1.24 \\
    \midrule
    \rowcolor{gray_bg}
    \textbf{Ours} & $640\times480$ & \textbf{3.09} & \textbf{4.25} & \underline{2.57} & \textbf{3.56} & \textbf{3.26} & 2.24 & \textbf{3.45} & \textbf{3.39} & \textbf{2.01} \\
    \bottomrule
    \end{tabular}
    }%
\end{table*}

\subsection{TDF3D-RoPE and Chain-of-Editing}

\begin{table*}[ht]
    \centering
    \small
    \setlength{\tabcolsep}{3pt}
    \caption{Ablation study on TDF3D-RoPE, which fuses inter-shot and intra-shot position encodings so that spatially corresponding patches across shots share zero relative position, enabling content-driven pixel-level correspondence. Evaluated on OpenVE-Bench with Gemini 2.5 Pro.}
    \label{tab:abl_rope}
    \resizebox{0.90\textwidth}{!}{%
    \begin{tabular}{c|c|c|cccccccc}
    \toprule
    TDF3D-RoPE & CoE & Overall & \begin{tabular}[c]{@{}c@{}}Global\\ Style\end{tabular} & \begin{tabular}[c]{@{}c@{}}Background\\ Change\end{tabular} & \begin{tabular}[c]{@{}c@{}}Local\\ Change\end{tabular} & \begin{tabular}[c]{@{}c@{}}Local\\ Remove\end{tabular} & \begin{tabular}[c]{@{}c@{}}Local\\ Add\end{tabular} & \begin{tabular}[c]{@{}c@{}}Subtitle\\ Edit\end{tabular} & \begin{tabular}[c]{@{}c@{}}Creative\\ Edit\end{tabular} & \begin{tabular}[c]{@{}c@{}}Camera\\ Edit\end{tabular} \\
    \midrule

     &  & 1.95 & 2.78 & 2.02 & 2.17 & 2.15 & 1.71 & 1.39 & 2.09 & 1.04 \\
     & \checkmark & 2.30 & 2.83 & 1.69 & 2.67 & 2.33 & 1.97 & 3.12 & 2.41 & 1.33 \\
     \checkmark &  & 2.53 & 3.06 & 1.94 & 2.88 & 2.85 & 1.97 & 3.49 & 2.69 & 1.29 \\
     \rowcolor{gray_bg}
    \checkmark & \checkmark & 2.65 & 3.64 & 2.15 & 3.03 & 3.11 & 1.82 & 3.20 & 2.89 & 1.22 \\

    \bottomrule
    \end{tabular}
    }%
\end{table*}

Tab.~\ref{tab:abl_rope} isolates the contributions of TDF3D-RoPE and CoE, revealing that spatial grounding is the more fundamental capability: TDF3D-RoPE alone yields a larger standalone gain than CoE alone, because establishing where to edit is a prerequisite for executing the edit faithfully.
The advantage of TDF3D-RoPE is sharpest on tasks that demand fine-grained localization (Subtitle Edit, Local Remove, and Local Change), where zero relative position between inter-shot patches resolves an otherwise ambiguous correspondence problem.
Without TDF3D-RoPE, the model must infer spatial alignment from content similarity alone, which is unreliable when foreground objects deform or when the editing instruction refers to a region not uniquely identifiable by appearance.
The two components are complementary but not fully additive: combining them lifts the overall score, yet the joint gain falls short of the sum of individual contributions, suggesting they partially address the same failure mode: the model misplacing edits when inter-shot context is weak.
An instructive exception is Subtitle Edit, where TDF3D-RoPE alone \emph{outperforms} the full model: the intermediate I$\to$I step in CoE introduces a degree of freedom that is unnecessary when TDF3D-RoPE already provides precise text-region anchoring, and the additional step adds noise rather than signal.
Background Change shows the reverse: CoE without TDF3D-RoPE actively \emph{hurts} performance, because the external image editor can freely alter large homogeneous regions that are semantically unconstrained, and without patch-level anchoring the video backbone has no mechanism to suppress these spurious changes.
Camera Edit remains consistently weak regardless of configuration, underscoring that neither positional encoding restructuring nor chain-based appearance editing addresses the core deficit: the model lacks geometric transformation capability.

\subsection{Temporal Position Randomization}

\begin{table*}[ht]
    \centering
    \small
    \setlength{\tabcolsep}{3pt}
    \caption{Ablation study on Temporal Position Randomization (TPR) for image editing training data, evaluated on OpenVE-Bench with Gemini 2.5 Pro.}
    \label{tab:abl_random_idx}
    \resizebox{0.90\textwidth}{!}{%
    \begin{tabular}{c|c|c|cccccccc}
    \toprule
    TPR & CoE & Overall & \begin{tabular}[c]{@{}c@{}}Global\\ Style\end{tabular} & \begin{tabular}[c]{@{}c@{}}Background\\ Change\end{tabular} & \begin{tabular}[c]{@{}c@{}}Local\\ Change\end{tabular} & \begin{tabular}[c]{@{}c@{}}Local\\ Remove\end{tabular} & \begin{tabular}[c]{@{}c@{}}Local\\ Add\end{tabular} & \begin{tabular}[c]{@{}c@{}}Subtitle\\ Edit\end{tabular} & \begin{tabular}[c]{@{}c@{}}Creative\\ Edit\end{tabular} & \begin{tabular}[c]{@{}c@{}}Camera\\ Edit\end{tabular} \\
    \midrule

     &  & 2.44 & 3.30 & 2.01 & 2.78 & 2.41 & 1.84 & 3.27 & 2.48 & 1.33 \\
     & \checkmark & 2.53 & 3.25 & 2.37 & 2.68 & 2.63 & 1.99 & 3.36 & 2.58 & 1.22 \\
     \checkmark &  & 2.53 & 3.06 & 1.94 & 2.88 & 2.85 & 1.97 & 3.49 & 2.69 & 1.29 \\
     \rowcolor{gray_bg}
    \checkmark & \checkmark & 2.65 & 3.64 & 2.15 & 3.03 & 3.11 & 1.82 & 3.20 & 2.89 & 1.22 \\

    \bottomrule
    \end{tabular}
    }%
\end{table*}

Tab.~\ref{tab:abl_random_idx} reveals that TPR is a prerequisite for CoE to work as intended, rather than an independent source of gain.
Without TPR, assigning a fixed temporal index to all training images creates a spurious correlation between position zero and the image modality (Section~\ref{sec:method}): the model learns to treat temporally-anchored tokens as modality identifiers rather than as spatial content references.
This has a concrete consequence at inference time: when CoE inserts the intermediate edited image into the context, the model partially ignores its appearance content and instead interprets it as a structural cue, so the spatial editing signal encoded in the keyframe fails to propagate faithfully to the generated video.
Categories demanding holistic appearance transfer, such as Global Style, are most sensitive to this failure mode: CoE without TPR actually \emph{degrades} performance there, because the anchor misleads more than it guides.
TPR resolves this by randomly assigning temporal positions to training images, forcing the model to ground correspondence in spatial content irrespective of when a frame nominally occurs.
Once this bias is removed, the edited keyframe serves as a genuine appearance anchor: the model can read the full $\mathbf{I}^s \to \mathbf{I}^t$ transformation and propagate it frame-by-frame, which is precisely the mechanism CoE is designed to exploit.
The interaction between the two components is therefore not additive but multiplicative: TPR prepares the representational substrate that makes CoE's test-time chain structurally meaningful, and tasks that require fine-grained spatial reasoning (Local Remove, Global Style) show the sharpest compounding benefit.

\subsection{Data Composition and Image Editing Quality}

Tab.~\ref{tab:abl_img_edit} shows a clear dissociation: PQ remains stable across configurations while SQ varies significantly, confirming that data composition governs instruction fidelity rather than low-level visual quality. V$\leftrightarrow$I$\leftrightarrow$I yields the strongest single-source SQ, as its chain structure forces an explicit semantic commitment before video generation. Adding V$\leftrightarrow$I$\leftrightarrow$I on top of I2I+V2V* produces the largest SQ gain, demonstrating that paired data and chain data play orthogonal roles.

\begin{table}[ht]
    \centering
    \small
    \setlength{\tabcolsep}{5pt}
    \caption{Ablation study on data composition evaluated on GEdit-Bench~\cite{geditbench}, with evaluation metrics including SQ (Semantic Consistency), PQ (Perceptual Quality), and O (Overall Score).}
    \label{tab:abl_img_edit}
    \resizebox{0.45\textwidth}{!}{%
    \begin{tabular}{ccc|c|c|c}
    \toprule
    I2I & V2V* & V$\leftrightarrow$I$\leftrightarrow$I & SQ & PQ & O \\
    \midrule
    \checkmark &  &  & 3.61 & 5.94 & 3.67 \\
     & \checkmark &  & 3.65 & 5.90 & 3.65 \\
     &  & \checkmark & 4.65 & 5.88 & 4.43 \\
    \checkmark & \checkmark &  & 4.54 & 5.82 & 4.42 \\
    \rowcolor{gray_bg}
    \checkmark & \checkmark & \checkmark & 5.23 & 5.66 & 4.86 \\
    \bottomrule
    \end{tabular}
    }%
\end{table}

\subsection{Comparison with Pure V2V Training}\label{sec:app:pure_v2v}

The V2V*-only rows of Tab.~\ref{tab:ablation} are trained on paired video data that is already reformatted into our decomposed, interleaved chain format.
To isolate the effect of decomposition and of the backbone itself, we train a \emph{Pure V2V} model with the same backbone on the same V2V data, but with no decomposition, no in-context formulation, and no CoE.
We compare it with OpenVE-Edit~\cite{openve3m2024}, which is trained on OpenVE's released V2V data, a strict superset of our pairwise video split (ours excludes the unreleased Creative Edit category).

\begin{table}[ht]
    \centering
    \small
    \setlength{\tabcolsep}{3pt}
    \caption{Comparison with Pure V2V training on OpenVE-Bench (Gemini 2.5 Pro). V2V: paired video editing data. Creat.: Creative Edit V2V data. Decomp.: decomposed, interleaved chain-of-editing format of V2V. IC: our V$\leftrightarrow$I$\leftrightarrow$I in-context data. I2I: OmniEdit image editing data.}
    \label{tab:pure_v2v}
    \resizebox{0.95\textwidth}{!}{%
    \begin{tabular}{l|ccccc|cccc}
    \toprule
    Method & V2V & Creat. & Decomp. & IC & I2I & Overall & Creative Edit & Global Style & Local Remove \\
    \midrule
    OpenVE-Edit & \checkmark & \checkmark & & & & 2.49 & 2.31 & 3.16 & 1.85 \\
    Pure V2V (ours) & \checkmark & & & & & 2.32 & 1.66 & 3.13 & 2.33 \\
    V2V* + CoE (ours) & \checkmark & & \checkmark & & & 2.57 & 1.88 & 3.32 & 2.87 \\
    \rowcolor{gray_bg}
    Full data + CoE (ours) & \checkmark & & \checkmark & \checkmark & \checkmark & \textbf{2.65} & \textbf{2.89} & \textbf{3.64} & \textbf{3.11} \\
    \bottomrule
    \end{tabular}
    }%
\end{table}

\begin{table}[ht]
    \centering
    \small
    \setlength{\tabcolsep}{4pt}
    \caption{Per-category gains of full data + CoE over Pure V2V on OpenVE-Bench (Gemini 2.5 Pro). Local Edit averages Local Change, Local Remove, and Local Add.}
    \label{tab:pure_v2v_delta}
    \resizebox{0.95\textwidth}{!}{%
    \begin{tabular}{l|c|cccc}
    \toprule
    Method & Overall & Global Style & Background Change & Local Edit & Creative Edit \\
    \midrule
    Pure V2V (w/o decomp., w/o CoE) & 2.32 & 3.13 & 1.89 & 2.22 & 1.66 \\
    Full data + CoE & 2.65 & 3.64 & 2.15 & 2.65 & 2.89 \\
    \midrule
    $\Delta$ & $+0.33$ ($+14.2\%$) & $+0.51$ ($+16.3\%$) & $+0.26$ ($+13.8\%$) & $+0.43$ ($+19.5\%$) & $+1.23$ ($+74.1\%$) \\
    \bottomrule
    \end{tabular}
    }%
\end{table}

With strictly less V2V data, Pure V2V underperforms OpenVE-Edit (2.32 vs.\ 2.49), confirming that our gains do not stem from a stronger backbone.
Applying decomposition and CoE to the same reduced data already surpasses it (2.57), and full data + CoE further improves both the overall score (2.65) and Creative Edit (2.89 vs.\ 2.31), despite having zero V2V supervision for that category.
As shown in Tab.~\ref{tab:pure_v2v_delta}, the overall gain over Pure V2V is $+14.2\%$, whereas the gain on Creative Edit is $+74.1\%$, roughly $5\times$ larger.
This gain comes from two sources, neither of which is paired video data: (1)~I2I and V$\leftrightarrow$I$\leftrightarrow$I data enabling implicit transfer via joint image--video modeling, and (2)~CoE inference providing explicit transfer via the in-context $\mathbf{I}^t$ demonstration.
Moreover, although I2I and V$\leftrightarrow$I$\leftrightarrow$I add 2.83M samples on top of V2V*, they increase the total number of visual training tokens by only $\sim$6\%, as image pairs are far cheaper than full videos; this yields a $14\%$ overall gain, a $>2\times$ efficiency ratio.
We therefore do not claim that video data is eliminated---the V$\to$I$\to$I$\to$V chain still requires video data to learn the final I$\to$V step---but that the dependence on large-scale \emph{paired} video editing data is substantially reduced, and we expect further scaling of image-level data to compound this improvement.

\subsection{Human Evaluation}\label{sec:app:human}

\mysubsubsec{Protocol.}
To verify that the MLLM-measured gap between V2V*-only + CoE (2.57) and full data + CoE (2.65) reflects a human-perceptible difference, we conduct a pairwise human evaluation over all 431 OpenVE-Bench samples.
We built a blinded side-by-side annotation tool: for each sample, the source video, the instruction, and the two edited outputs are shown as anonymized ``Video 1'' and ``Video 2''; left/right placement is randomized per sample to remove position bias, and the order of samples is shuffled.
For each of three dimensions (Prompt Following, Consistency, and Visual Quality), the rater selects ``Video 1 better'', ``Video 2 better'', or ``Tie''.
Ten independent raters each annotated all 431 samples.
We will release the annotation tool and the raw per-clip, per-rater votes.

\begin{table}[ht]
    \centering
    \small
    \begin{minipage}[t]{0.56\linewidth}
    \centering
    \caption{Pairwise human evaluation on OpenVE-Bench (431 samples). A: V2V*-only + CoE; B: full data + CoE.}
    \label{tab:human_eval}
    \begin{tabular}{l|ccc}
    \toprule
    Dimension & A wins & B wins & Tie \\
    \midrule
    Prompt Following & 24\% & \textbf{38\%} & 39\% \\
    Consistency & 9\% & \textbf{28\%} & 63\% \\
    Visual Quality & 14\% & \textbf{15\%} & 71\% \\
    \bottomrule
    \end{tabular}
    \end{minipage}
    \hfill
    \begin{minipage}[t]{0.40\linewidth}
    \centering
    \caption{Inter-rater agreement (10 raters), aggregated over all three dimensions.}
    \label{tab:human_agree}
    \begin{tabular}{l|c}
    \toprule
    Fleiss' $\kappa$~\cite{fleiss1971kappa} & 0.721 \\
    Unanimous rate (10/10) & 59.3\% \\
    Mean pairwise agreement & 90.4\% \\
    \bottomrule
    \end{tabular}
    \end{minipage}
\end{table}

\mysubsubsec{Results and agreement.}
As shown in Tab.~\ref{tab:human_eval}, B wins more often than A on every dimension, most clearly on Prompt Following and Consistency, in line with the MLLM scores.
Since OpenVE-Bench uses a 1--5 rating scale, the absolute range is inherently narrow, and a 0.08-point gap should be read in that context.
Fleiss' $\kappa = 0.721$ falls in the ``substantial agreement'' band (0.61--0.80)~\cite{landis1977kappa}, and the 90.4\% mean pairwise agreement indicates that almost any two raters agree on the vast majority of items.
The unanimous rate is lower (59.3\%), as expected: the boundary between ``Tie'' and a narrow preference is soft, and 10-way unanimity is a much stricter bar than pairwise agreement.

\mysubsubsec{Why ties dominate Visual Quality and Consistency.}
(1)~Visual Quality depends heavily on the foundation model, which is shared identically between the two configurations (they differ only in training data composition), so raters have little basis to prefer either.
(2)~For Consistency, V2V*-only + CoE more often fails to execute the requested edit, and a failed or partial edit trivially leaves unrelated regions unchanged, which reads as good consistency even though the edit did not succeed.
(3)~More generally, whether an instruction was followed is usually an unambiguous judgment, whereas subtle background drift or per-frame artifacts are harder to perceive and compare, so raters default to ``Tie'' more often on Consistency and Visual Quality than on Prompt Following.

\subsection{Evaluation with Objective VBench Metrics}\label{sec:app:vbench}

\mysubsubsec{Motivation.}
All benchmark scores in the main paper are produced by MLLM judges, which may share systematic biases. Although the rank ordering is consistent across two independent MLLM judges (Appendix~\ref{sec:app:seedvl}), we further seek evidence that does not rely on any MLLM. We therefore evaluate with VBench~\cite{huang2024vbench}, a widely used benchmark suite for video generative models whose dimensions are computed by dedicated perception models and hand-crafted measures rather than by an MLLM.

\mysubsubsec{Metrics.}
We report two VBench quality metrics: \texttt{imaging\_quality}, which measures per-frame low-level distortions such as over-exposure, noise, and blur using the MUSIQ image quality predictor, and \texttt{quality\_avg}, which aggregates VBench's video-quality dimensions to reflect the overall visual and temporal quality of the generated video. Both are computed on the edited videos that each method produces for OpenVE-Bench; higher is better.

\mysubsubsec{Results.}
As shown in Tab.~\ref{tab:vbench}, \ours achieves the best scores on both metrics among recent baselines. The margin is clearest on \texttt{imaging\_quality} ($+0.027$ over the strongest baseline, ICVE), indicating that routing the edit through an explicit keyframe does not degrade per-frame fidelity, while \texttt{quality\_avg} remains on par with or slightly above DITTO.

\mysubsubsec{Scope.}
These metrics assess only the visual and temporal quality of the output; they do not measure whether the edit follows the instruction or preserves unedited content. They are therefore complementary to, rather than a replacement for, the MLLM-based evaluation, which explicitly scores instruction compliance and consistency.

\begin{table}[ht]
    \centering
    \small
    \caption{VBench objective metrics on OpenVE-Bench.}
    \label{tab:vbench}
    \begin{tabular}{l|cc}
    \toprule
    Method & imaging\_quality & quality\_avg \\
    \midrule
    Lucy-Edit~\cite{lucyedit2024} & 0.652 & 0.800 \\
    ICVE~\cite{icve2024} & 0.688 & 0.812 \\
    DITTO~\cite{ditto2024} & 0.687 & 0.816 \\
    \rowcolor{gray_bg}
    \textbf{Ours} & \textbf{0.715} & \textbf{0.819} \\
    \bottomrule
    \end{tabular}
\end{table}

\subsection{Test-time Scaling and Efficiency}\label{sec:app:tts}

\begin{figure}[ht]
    \centering
    \includegraphics[width=0.48\linewidth]{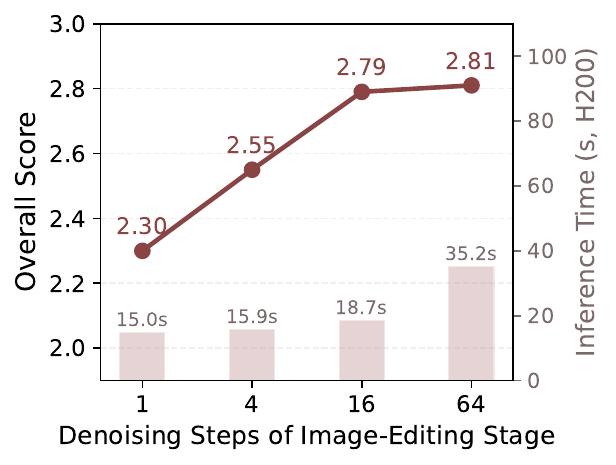}
    \caption{Test-time scaling of CoE: overall score (line, left axis) and inference time (bars, right axis) as the number of denoising steps of the intermediate image-editing stage increases. Scored by Gemini 3.1 Flash Lite.}
    \label{fig:tts_steps}
\end{figure}

\mysubsubsec{Test-time scaling.}
We vary the number of denoising steps of the intermediate image-editing stage of CoE (1, 4, 16, 64), holding everything else fixed, and evaluate on the full OpenVE-Bench (Tab.~\ref{tab:tts}, Fig.~\ref{fig:tts_steps}).
The overall score improves monotonically from 2.30 to 2.81 ($+22.2\%$ relative), \ie, additional test-time compute yields consistent quality gains without retraining, and the trend holds across the reported editing categories.
These results are scored with Gemini 3.1 Flash Lite, since Gemini 2.5 Pro was temporarily inaccessible during this experiment; absolute scores are therefore not directly comparable with the Gemini 2.5 Pro numbers elsewhere in the paper, but the judge is held fixed across all four settings, so the trend itself is valid.

\begin{table}[ht]
    \centering
    \small
    \setlength{\tabcolsep}{5pt}
    \caption{Test-time scaling via the number of denoising steps of the intermediate image-editing stage, evaluated on OpenVE-Bench by Gemini 3.1 Flash Lite. Inference time is measured on one H200 GPU.}
    \label{tab:tts}
    \resizebox{0.95\textwidth}{!}{%
    \begin{tabular}{c|c|c|ccccccc}
    \toprule
    Steps & Time & Overall & \begin{tabular}[c]{@{}c@{}}Background\\ Change\end{tabular} & \begin{tabular}[c]{@{}c@{}}Creative\\ Edit\end{tabular} & \begin{tabular}[c]{@{}c@{}}Global\\ Style\end{tabular} & \begin{tabular}[c]{@{}c@{}}Local\\ Add\end{tabular} & \begin{tabular}[c]{@{}c@{}}Local\\ Change\end{tabular} & \begin{tabular}[c]{@{}c@{}}Local\\ Remove\end{tabular} & \begin{tabular}[c]{@{}c@{}}Subtitle\\ Edit\end{tabular} \\
    \midrule
    1 & 15.02s & 2.30 & 1.72 & 2.03 & 3.59 & 1.64 & 2.48 & 2.44 & 3.03 \\
    4 & 15.93s & 2.55 & 2.29 & 2.49 & 3.60 & 1.80 & 2.61 & 2.85 & 3.28 \\
    16 & 18.69s & 2.79 & 2.69 & 2.72 & \textbf{3.67} & \textbf{2.24} & 2.96 & \textbf{3.15} & 3.42 \\
    64 & 35.20s & \textbf{2.81} & \textbf{2.76} & \textbf{2.97} & 3.55 & 2.10 & \textbf{3.05} & 3.06 & \textbf{3.57} \\
    \bottomrule
    \end{tabular}
    }%
\end{table}

\mysubsubsec{Inference latency.}\label{sec:app:latency}
Since CoE consists of two sequential stages, we measure its inference latency against two baselines from Tab.~\ref{tab:perf_gemini} (ICVE and Lucy-Edit) and AnyV2V, all on the same single H200 GPU with 50 denoising steps (Tab.~\ref{tab:latency}).
Despite the two-stage design, \ours is faster than ICVE, comparable to Lucy-Edit, and over an order of magnitude faster than AnyV2V, so the explicit intermediate edit does not come at a large cost to practical usability.

\begin{table}[ht]
    \centering
    \small
    \caption{Inference latency on one H200 GPU (50 denoising steps for all methods).}
    \label{tab:latency}
    \begin{tabular}{l|c}
    \toprule
    Method & Inference Time \\
    \midrule
    AnyV2V~\cite{ku2024anyv2v} & 11min 46s \\
    ICVE~\cite{icve2024} & 118.1s \\
    Lucy-Edit~\cite{lucyedit2024} & 39.8s \\
    \rowcolor{gray_bg}
    Ours (Stage 1 + Stage 2) & 18.4s + 35.1s = 53.5s \\
    \bottomrule
    \end{tabular}
\end{table}
\subsection{Comparison with AnyV2V}\label{sec:app:anyv2v}

AnyV2V~\cite{ku2024anyv2v} is closely related to our I$\to$V design, as it also leverages image editing for video editing.
We evaluate it on the full OpenVE-Bench and report the results in Tab.~\ref{tab:anyv2v}.
These scores are produced by Gemini 3.0 Flash Lite because Gemini 2.5 Pro was inaccessible at the time, so they are not directly comparable with Tab.~\ref{tab:perf_gemini}.
Both of our checkpoints outperform AnyV2V by a large margin, and AnyV2V is also over $13\times$ slower (Appendix~\ref{sec:app:latency}).

\begin{table}[ht]
    \centering
    \small
    \setlength{\tabcolsep}{3pt}
    \caption{Comparison with AnyV2V on OpenVE-Bench, evaluated by Gemini 3.0 Flash Lite.}
    \label{tab:anyv2v}
    \resizebox{0.95\textwidth}{!}{%
    \begin{tabular}{l|c|c|cccccccc}
    \toprule
    Method & \#Reso. & Overall & \begin{tabular}[c]{@{}c@{}}Background\\ Change\end{tabular} & \begin{tabular}[c]{@{}c@{}}Camera\\ Edit\end{tabular} & \begin{tabular}[c]{@{}c@{}}Creative\\ Edit\end{tabular} & \begin{tabular}[c]{@{}c@{}}Global\\ Style\end{tabular} & \begin{tabular}[c]{@{}c@{}}Local\\ Add\end{tabular} & \begin{tabular}[c]{@{}c@{}}Local\\ Change\end{tabular} & \begin{tabular}[c]{@{}c@{}}Local\\ Remove\end{tabular} & \begin{tabular}[c]{@{}c@{}}Subtitle\\ Edit\end{tabular} \\
    \midrule
    AnyV2V~\cite{ku2024anyv2v} & $360\times640$ & 1.62 & 1.38 & 1.16 & 2.17 & 3.05 & 1.29 & 1.56 & 1.16 & 1.37 \\
    Ours (Stage 1) & $256\times256$ & 2.79 & 2.71 & 1.15 & 2.88 & 3.52 & 2.33 & 2.96 & 3.17 & 3.36 \\
    \rowcolor{gray_bg}
    Ours (Stage 2) & $480\times640$ & \textbf{3.26} & \textbf{3.16} & \textbf{2.26} & \textbf{3.76} & \textbf{3.95} & \textbf{2.55} & \textbf{3.62} & \textbf{3.45} & \textbf{3.37} \\
    \bottomrule
    \end{tabular}
    }%
\end{table}

\section{More Qualitative Results}\label{sec:app:qualitative}

\begin{figure*}[t]
    \centering
    \includegraphics[width=\textwidth]{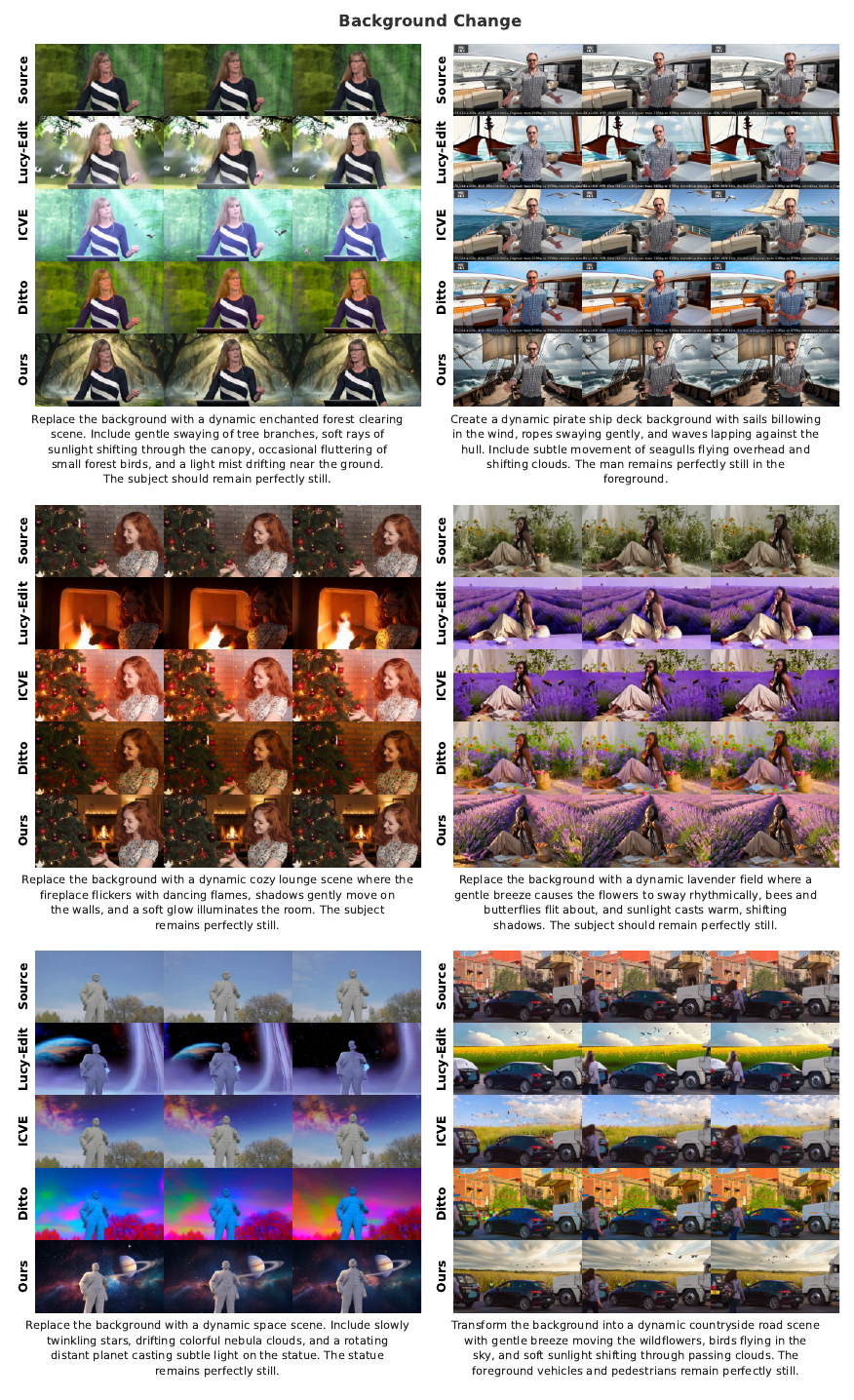}
    \caption{Qualitative comparison on \textbf{background change} editing.}
    \label{fig:app:background_change}
\end{figure*}

\begin{figure*}[t]
    \centering
    \includegraphics[width=\textwidth]{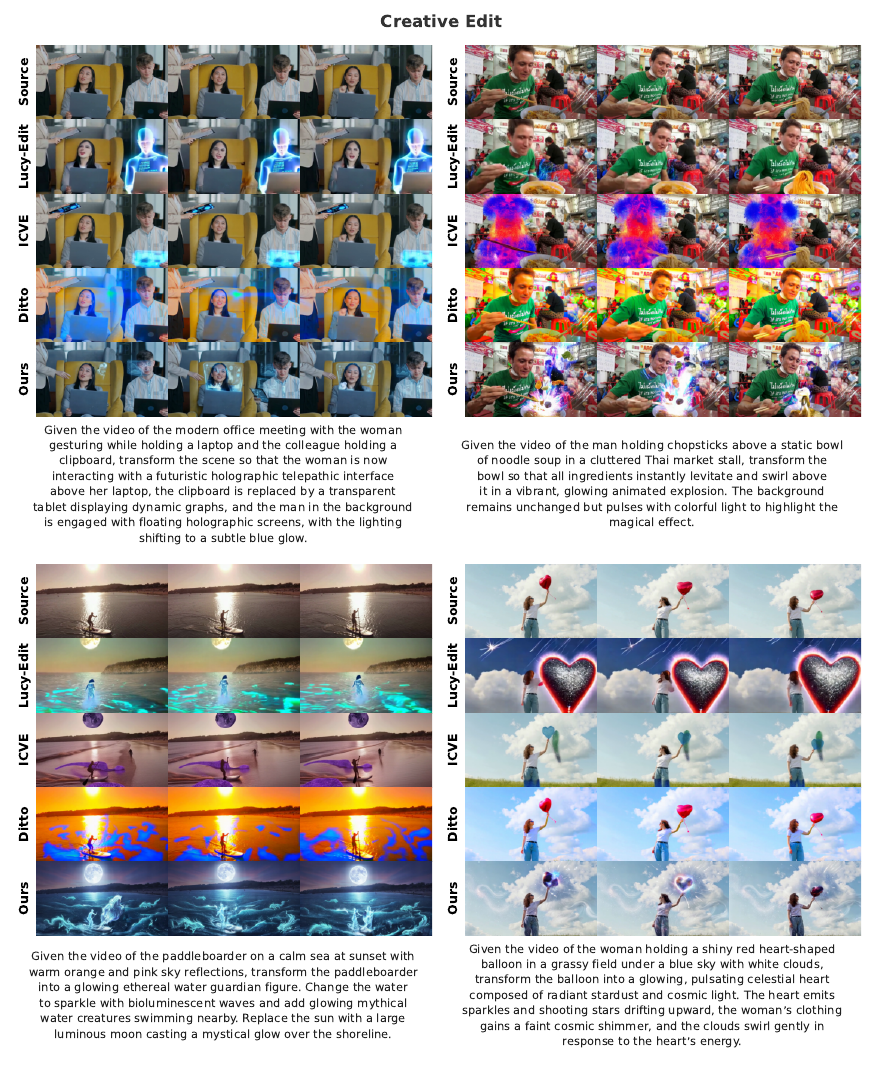}
    \caption{Qualitative comparison on \textbf{creative edit} editing.}
    \label{fig:app:creative_edit}
\end{figure*}

\begin{figure*}[t]
    \centering
    \includegraphics[width=\textwidth]{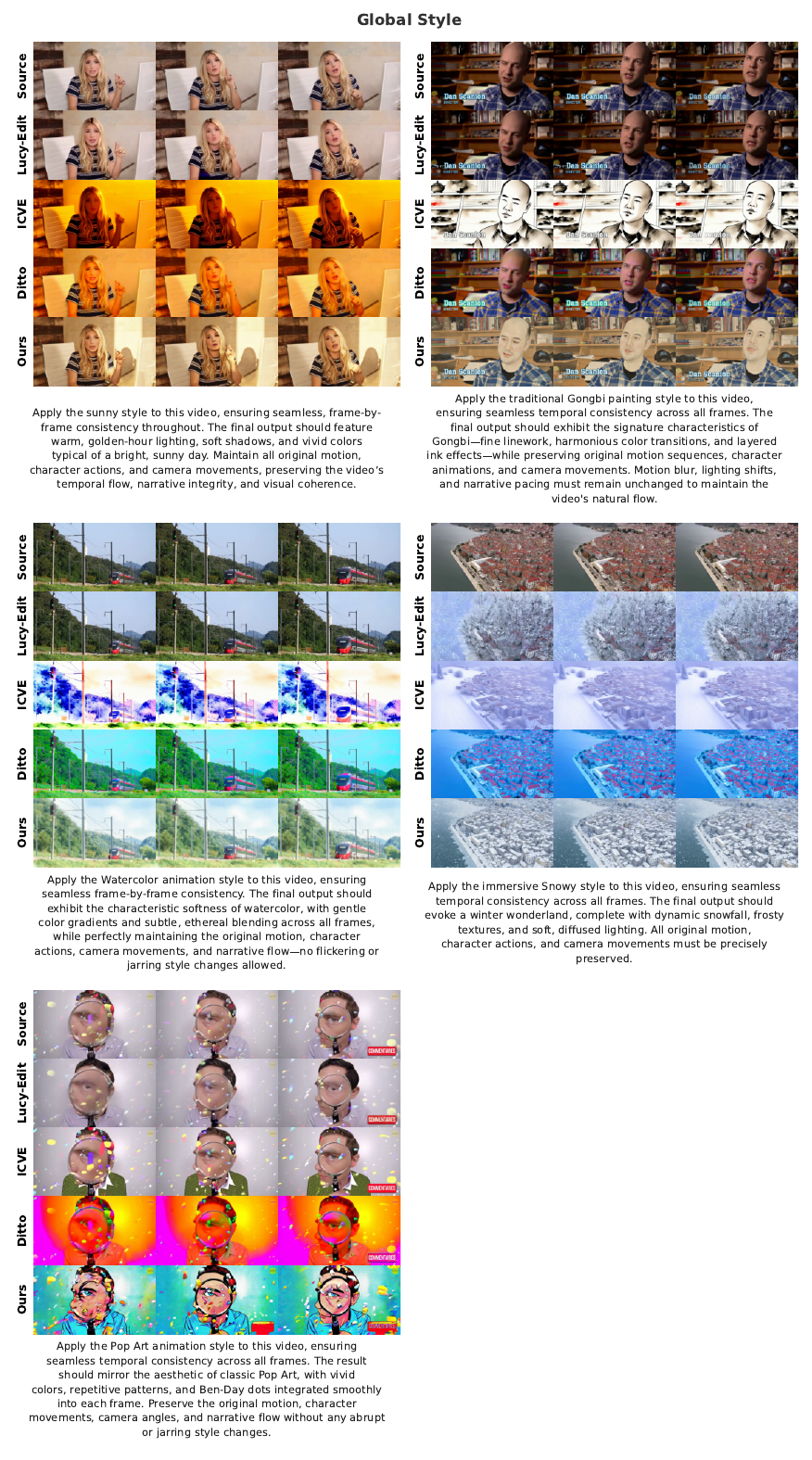}
    \caption{Qualitative comparison on \textbf{global style} editing.}
    \label{fig:app:global_style}
\end{figure*}

\begin{figure*}[t]
    \centering
    \includegraphics[width=\textwidth]{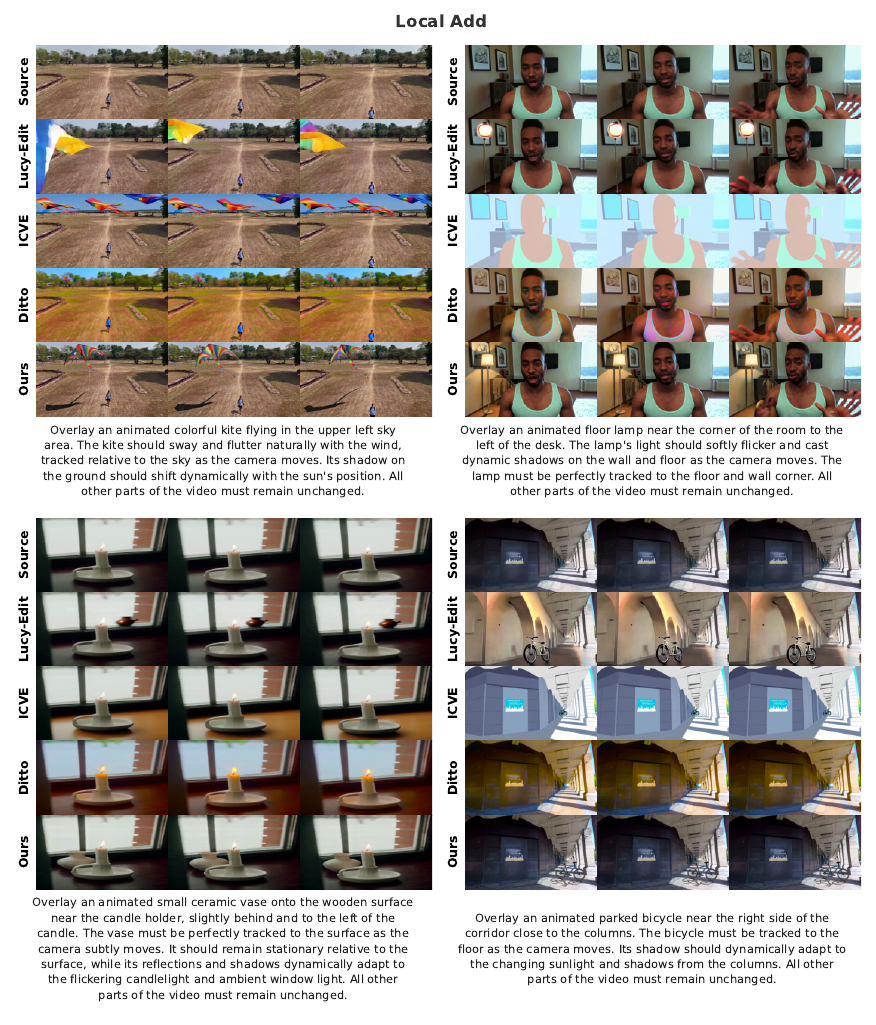}
    \caption{Qualitative comparison on \textbf{local add} editing.}
    \label{fig:app:local_add}
\end{figure*}

\begin{figure*}[t]
    \centering
    \includegraphics[width=\textwidth]{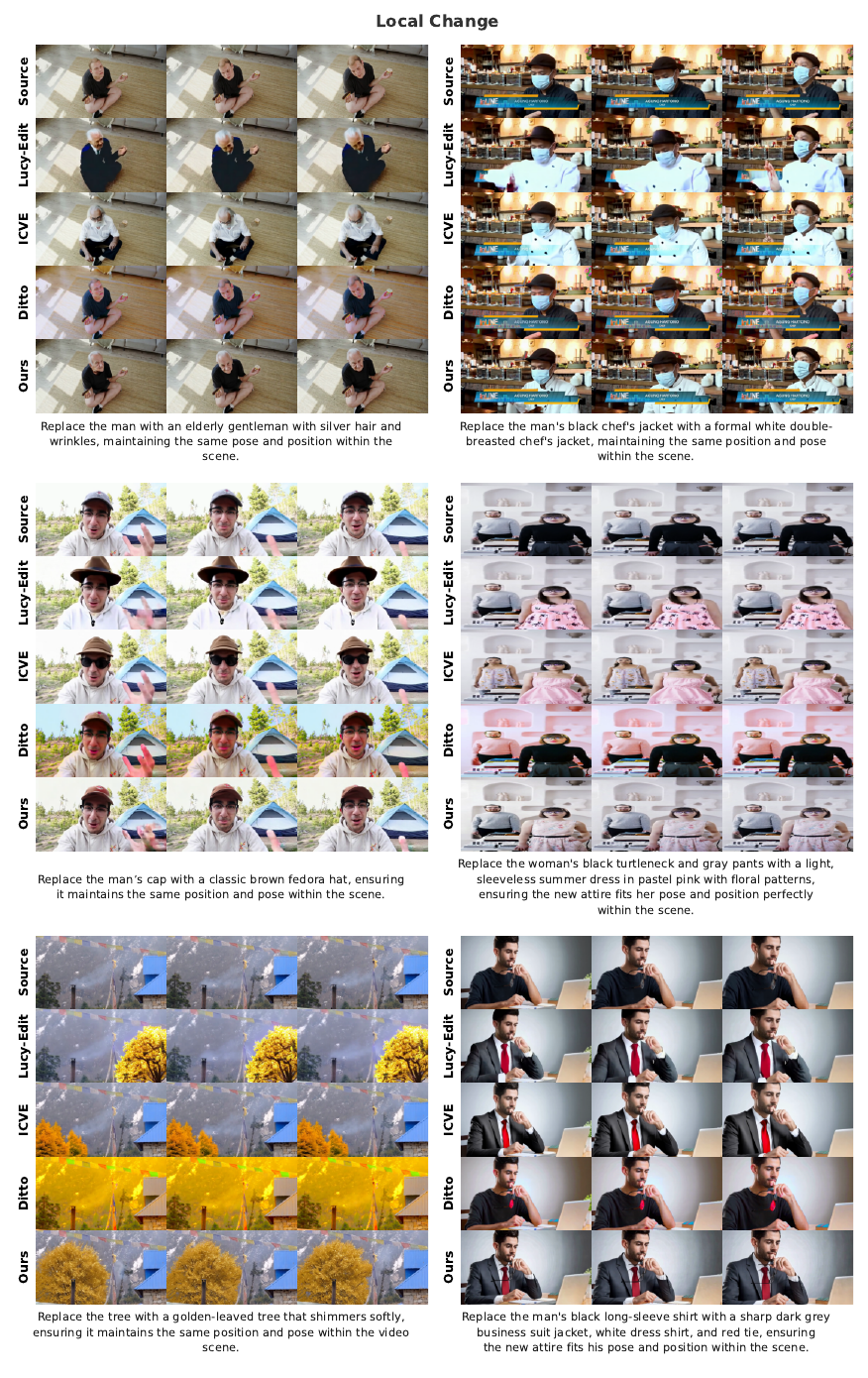}
    \caption{Qualitative comparison on \textbf{local change} editing.}
    \label{fig:app:local_change}
\end{figure*}

\begin{figure*}[t]
    \centering
    \includegraphics[width=\textwidth]{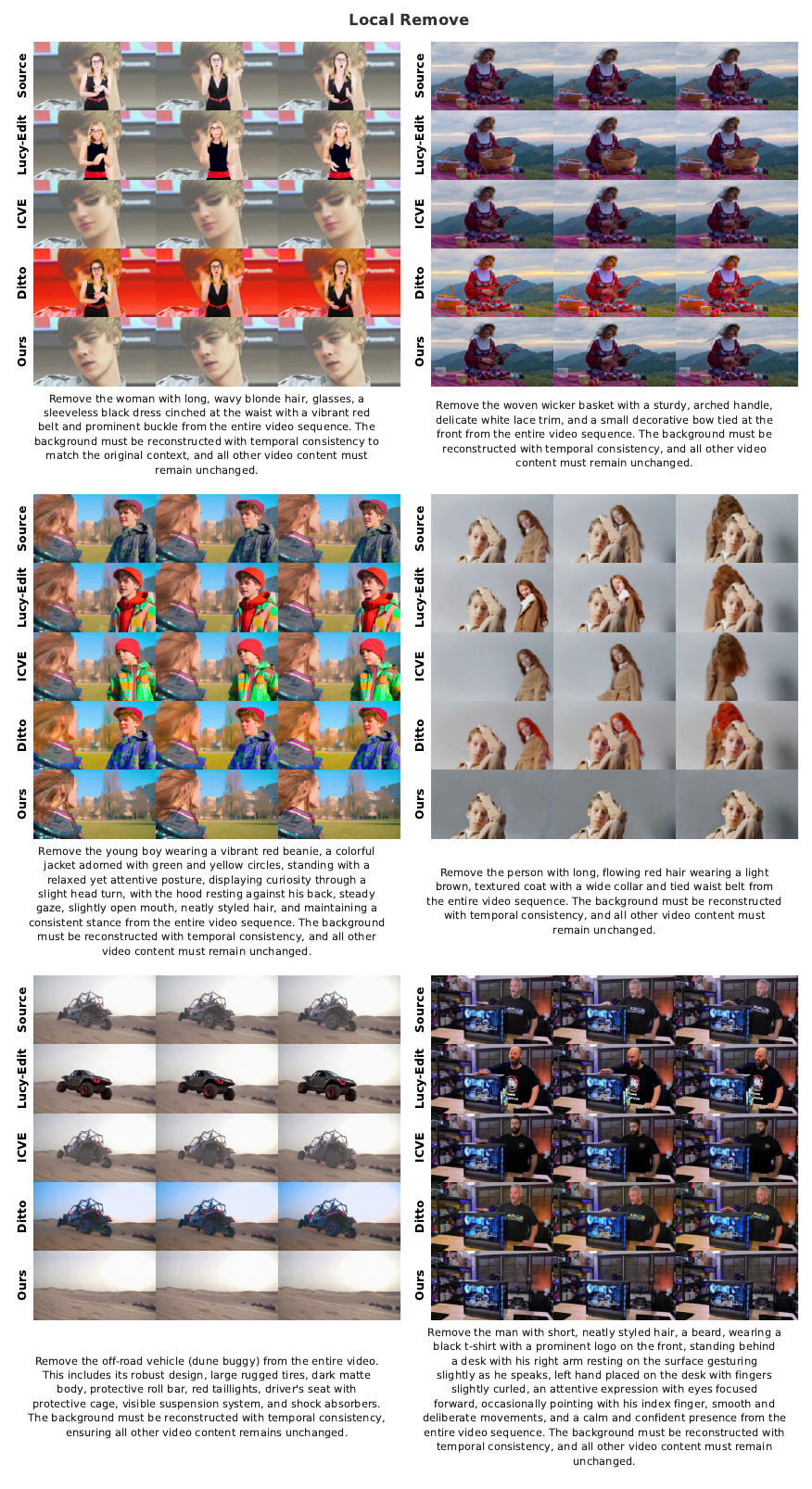}
    \caption{Qualitative comparison on \textbf{local remove} editing.}
    \label{fig:app:local_remove}
\end{figure*}

\begin{figure*}[t]
    \centering
    \includegraphics[width=\textwidth]{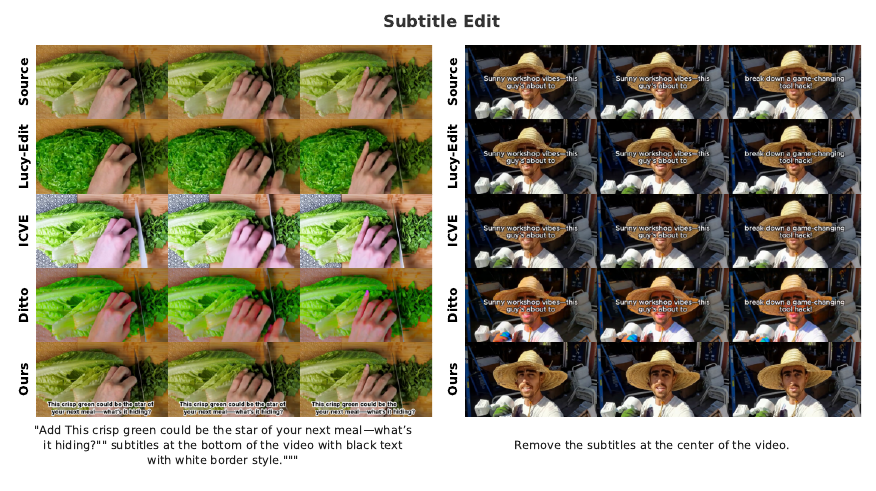}
    \caption{Qualitative comparison on \textbf{subtitle edit} editing.}
    \label{fig:app:subtitle_edit}
\end{figure*}

This section provides extended qualitative results complementing the main paper.
Figs.~\ref{fig:app:background_change}--\ref{fig:app:subtitle_edit} compare \ours against all open-source baselines across seven editing categories.
Fig.~\ref{fig:app:coe_variants} then isolates the contribution of Chain-of-Editing by contrasting V2V direct inference with CoE using self-editing and an external editor.

\subsection{Qualitative Comparison between Baselines and Ours}

\noindent
Each figure displays the source video, the editing instruction, and the outputs of all seven open-source baselines alongside \ours.
For spatially-aligned categories (Background Change, Global Style, Local Add, Local Change, Local Remove, Subtitle Edit), \ours consistently produces edits that precisely follow the instruction while preserving the identity of untouched regions and maintaining temporal coherence across frames.
Baseline methods tend to fail in one of two modes: either the edit is not applied at all (instruction non-compliance), or the edit bleeds into unrelated regions and introduces flicker.
For the non-spatially-aligned Creative Edit category, most baselines struggle with semantic reinterpretation; \ours benefits from routing the edit through the image domain, which provides a concrete appearance blueprint before video generation begins.

\subsection{Qualitative Comparison between V2V and CoE Variants of VINCIE-NExT}

\begin{figure*}[t]
    \centering
    \includegraphics[width=\textwidth]{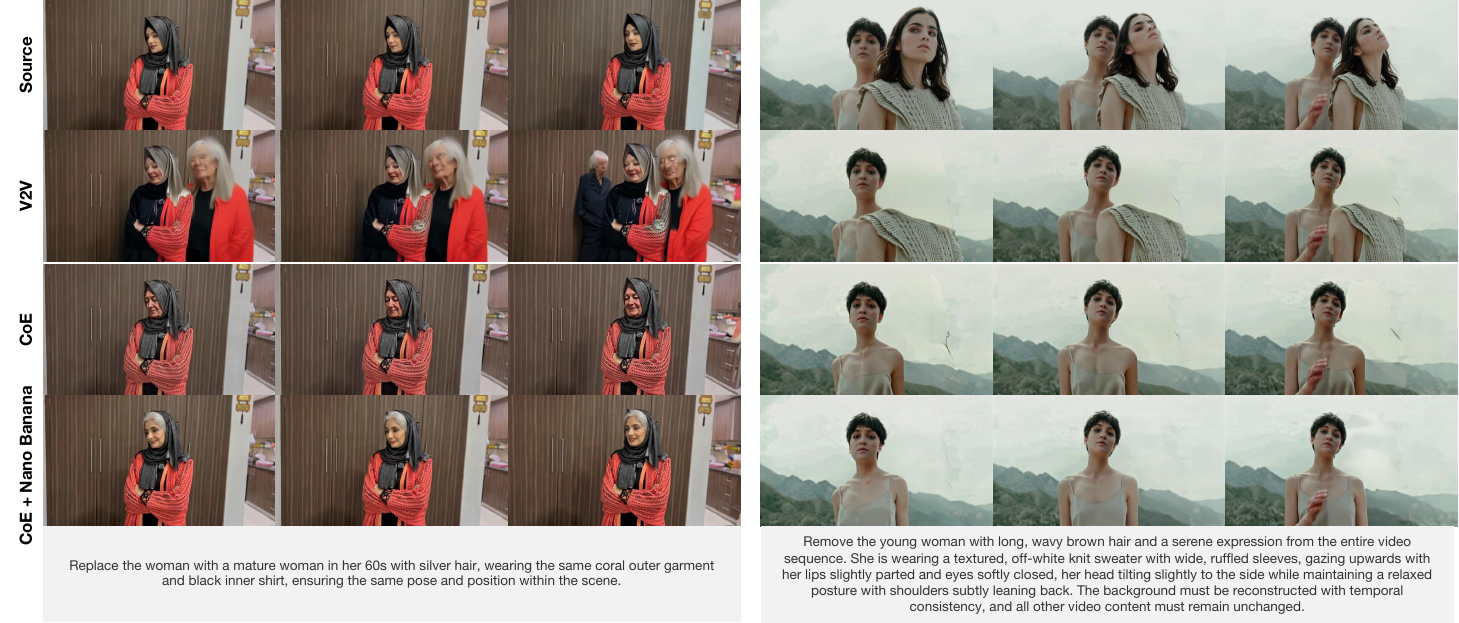}
    \caption{Qualitative comparison of three inference modes of \ours on subject replacement (\textbf{left}) and subject removal (\textbf{right}): V2V direct inference, CoE with self-editing, and CoE with an external image editor.}
    \label{fig:app:coe_variants}
\end{figure*}

Fig.~\ref{fig:app:coe_variants} isolates the effect of Chain-of-Editing on two challenging tasks: subject replacement and subject removal.
V2V direct inference, while capable of approximate edits, produces temporal flickering and incomplete object removal because the model must simultaneously infer \emph{what} to change and generate consistent video dynamics.
CoE with self-editing addresses this by committing to an edited keyframe first: the intermediate image explicitly resolves the semantic ambiguity before the video backbone generates the remaining frames, yielding sharper boundaries and more stable motion.
Substituting the self-editor with a stronger external image editor further improves semantic accuracy---particularly for subject replacement, where the external model can synthesize convincing appearance details that the self-editor occasionally blurs.
Crucially, this substitution requires no retraining: the modular I$\to$I interface means that improvements in image editing transfer directly to video editing, demonstrating a clear path for future scaling.

\section*{Future Work}

Two directions are promising for extending \ours.
First, a more unified model could span image, video, and other modalities (\eg, audio and 3D) and couple understanding with generation in a single backbone~\cite{sun2024emu2,xiao2024omnigen,omnivideo2024}, allowing it to reason about the source video before deciding how to edit it rather than following a fixed V$\to$I$\to$I$\to$V decomposition.
Second, Chain-of-Editing can be generalized into an agentic framework that plans multi-step edits, invokes external tools (\eg, segmentation, image editors, and quality verifiers) on demand, and leverages retrieval-augmented generation~\cite{qu2021dynamic,qu2025tiger} to ground edits involving rare concepts or novel content.

\section*{Limitations}

Despite promising results, \ours has several limitations.
First, output quality is bottlenecked by the intermediate I$\to$I step: errors in the edited keyframe (\eg, hallucinated details or incomplete edits) propagate to every output frame.
As shown in Fig.~\ref{fig:abl_coe_qwen_nano}, self-editing underperforms on edits requiring geometric transformation or novel-content synthesis, while stronger external editors consistently improve video quality.
Since Stage~2 treats $\mathbf{I}^t$ as a hard spatial anchor, a badly mismatched edit will propagate rather than be ``corrected''---which is the intended behavior for faithfully preserving the $\mathbf{I}^s{\to}\mathbf{I}^t$ edit. This risk is mitigated by image editing being a far more mature task than video editing; we leave a systematic stress test to future work.
Second, a single keyframe may not adequately characterize videos with significant temporal variation or large camera motion.
Third, Chain-of-Editing requires two sequential diffusion stages; although its latency (53.5s on one H200) is lower than ICVE and comparable to Lucy-Edit (Appendix~\ref{sec:app:latency}), longer test-time chains add cost.
Finally, evaluation is limited to OpenVE-Bench; generalization to long or highly dynamic videos remains open.

\section*{Scope}

\ours is designed for instruction-driven appearance editing of short video clips, including style transfer, object attribute modification, and background replacement.
It is not designed for structural edits that require inserting or removing objects across frames, nor for edits that depend on physical simulation or long-range temporal reasoning.
The framework assumes a static or slowly varying camera; videos with rapid cuts or severe occlusion may challenge the keyframe-grounded editing paradigm.
User-supplied image demonstrations extend the framework beyond text instructions, but crafting effective demonstrations for highly complex edits remains non-trivial.

\section*{Broader Impact}

\ours lowers the barrier to high-quality video editing by substantially reducing the need for large-scale paired video data and expensive per-video fine-tuning, making capable editing tools more accessible to individual creators and small studios.
At the same time, the same capability can be misused to produce deceptive or non-consensual manipulations of video content.
We encourage the community to pair advances in generative editing with robust provenance and detection tools, and we endorse established responsible-disclosure norms when deploying such systems in consumer-facing products.

\end{document}